%% file: d2s_eccv18.tex
\RequirePackage{amsmath}
\documentclass[runningheads]{llncs}
\usepackage{graphicx}
\usepackage{amssymb} 
\usepackage{color}

\usepackage{epsfig}
\usepackage[pagebackref=true,breaklinks=true,colorlinks,bookmarks=false]{hyperref}

\usepackage{pgfplots}
\usepackage{pgf,tikz}
\usepackage{mathrsfs}
\usepackage{xspace}
\pgfplotsset{compat=1.5}
\usepackage[caption=false]{subfig}

\newcommand{\figref}[1]{\mbox{Fig.~\ref{#1}}}
\newcommand{\tabref}[1]{\mbox{Table~\ref{#1}}}

\definecolor{colorAqua}{RGB}{75,172,198}
\definecolor{colorGreen}{RGB}{155,187,89}
\definecolor{colorViolet}{RGB}{112,48,160}
\definecolor{colorOrange}{RGB}{247,150,70}
\definecolor{colorRed}{RGB}{192,80,77}
\definecolor{colorBlue}{RGB}{79,129,189}
\definecolor{colorYellow}{RGB}{255,192,0}
\def\addlegendimage{\csname pgfplots@addlegendimage\endcsname}
\def\addlegendentry{\csname pgfplots@addlegendentry\endcsname}

\makeatletter
\DeclareRobustCommand\onedot{\futurelet\@let@token\@onedot}
\def\@onedot{\ifx\@let@token.\else.\null\fi\xspace}

\def\etal{\emph{et al}\onedot}
\makeatother

\begin{document}
\title{MVTec D2S: Densely Segmented Supermarket Dataset}

\titlerunning{MVTec D2S: Densely Segmented Supermarket Dataset}
%
\author{Patrick Follmann \inst{1,2}\orcidID{0000-0001-5400-2384} \and 
Tobias B\"ottger \inst{1,2}\orcidID{0000-0002-5404-8662} \and 
Philipp H\"artinger \inst{1}\orcidID{0000-0002-7093-6280} \and 
Rebecca K\"onig \inst{1}  \orcidID{0000-0002-4169-6759} \and
Markus Ulrich \inst{1}\orcidID{0000-0001-8457-5554}}
%
\authorrunning{Follmann \etal}
%

\institute{MVTec Software GmbH, 80634 Munich, Germany
\email{\{follmann,boettger,haertinger,koenig,ulrich\}@mvtec.com}\\
\url{https://www.mvtec.com/research} \and
Technical University of Munich, 80333 Munich, Germany}

  \makeatletter
\let\@oldmaketitle\@maketitle
\renewcommand{\@maketitle}{\@oldmaketitle
  \setlength{\fboxsep}{0pt}
  \hfill
  \mbox{
    \includegraphics[width=0.23\textwidth]{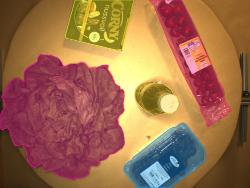}
    \includegraphics[width=0.23\textwidth]{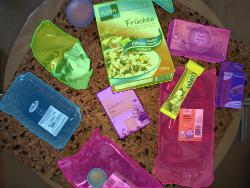}
    \includegraphics[width=0.23\textwidth]{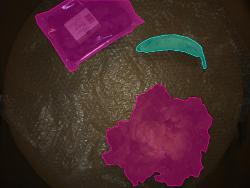}
    \includegraphics[width=0.23\textwidth]{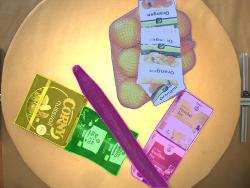}
  }
  \hfill
}
\makeatother

\maketitle              
\begin{abstract}
  We introduce the Densely Segmented Supermarket (D2S) dataset, a novel benchmark
  for instance-aware semantic segmentation in an industrial domain. It contains
  21\,000 high-resolution images with pixel-wise labels of all object instances.
  The objects comprise groceries and everyday products from 60 categories. The
  benchmark is designed such that it resembles the real-world setting of an
  automatic checkout, inventory, or warehouse system. The training images only
  contain objects of a single class on a homogeneous background, while the
  validation and test sets are much more complex and diverse. To further
  benchmark the robustness of instance segmentation methods, the scenes are
  acquired with different lightings, rotations, and backgrounds. We ensure that
  there are no ambiguities in the labels and that every instance is labeled
  comprehensively. The annotations are pixel-precise and allow using crops of
  single instances for articial data augmentation. The dataset covers several
  challenges highly relevant in the field, such as a limited amount of training
  data and a high diversity in the test and validation sets. The evaluation of
  state-of-the-art object detection and instance segmentation methods on D2S
  reveals significant room for improvement.
  \keywords{instance segmentation dataset, industrial application}
\end{abstract}

\input{section_intro}

\input{section_dataset}

\input{section_splits}

\input{section_statistics}

\input{section_benchmark}

\input{section_conclusion}

\section*{Acknowledgements}

We want to thank the students Clarissa Siegfarth, Bela Jugel, Thomas Beraneck, Johannes K\"ohne, Christoph Ziegler and Bernie St\"offler for their help to acquire and annotate the dataset.

\clearpage

\appendix
\section*{Appendix}
\section{Intra- and Inter-class Visual Variations}

\begin{figure}
  \centering
  \scriptsize
  \setlength{\tabcolsep}{0.5pt}
  \begin{tabular}{ccccc}
    \includegraphics[width=0.19\textwidth]{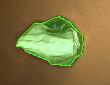} &
    \includegraphics[width=0.19\textwidth]{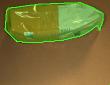} &
    \includegraphics[width=0.19\textwidth]{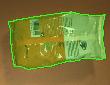} &
    \includegraphics[width=0.19\textwidth]{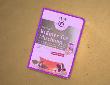} &
    \includegraphics[width=0.19\textwidth]{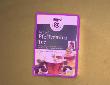}
  \end{tabular}
  \begin{tabular}{cccccc}
    \includegraphics[width=0.155\textwidth]{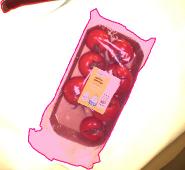} &
    \includegraphics[width=0.155\textwidth]{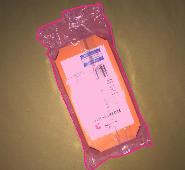} &
    \includegraphics[width=0.155\textwidth]{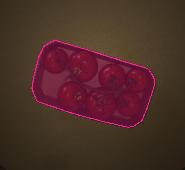} &
    \includegraphics[width=0.155\textwidth]{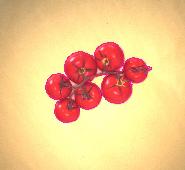} &
    \includegraphics[width=0.155\textwidth]{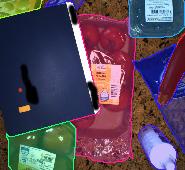} &
    \includegraphics[width=0.155\textwidth]{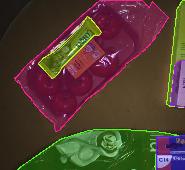} \\
  \end{tabular}
  \caption{{\bf Intra- and inter-class variability}
    A collection of exemplary objects with annotations that show the variety of transformations and deformations, as well as the intra-class variability e.g. with respect to packaging of an object. Also note the small difference between the two different classes of tea in row one on the right.}
  \label{fig:class_variability}
\end{figure}

We show some of the intra-class visual variations in \figref{fig:class_variability}. Due to the rotations and different lightings, the objects appear with various reflections. Furthermore, many different rigid transformations have been applied. Further variations are different degrees of partial occlusions. For objects like the cocoa package in the top left of \figref{fig:class_variability} even non-rigid deformations are contained. In comparison to other datasets, some of our classes have only subtle appearance differences, adding the challenge of fine-grained reasoning to D2S. 

\section{Qualitative Results}
\begin{figure}
  \centering
  \scriptsize
  \setlength{\tabcolsep}{0.5pt}
  \begin{tabular}{ccccccc}
    \textbf{ground truth} &
    \texttt{rot0\_light0} &
    \texttt{rot0} &
    \texttt{light0} &
    \texttt{train} &
    \texttt{aug} &
    \texttt{train+aug}
    \\
    \includegraphics[width=0.14\textwidth]{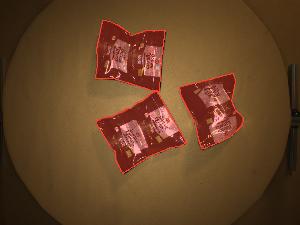} &
    \includegraphics[width=0.14\textwidth]{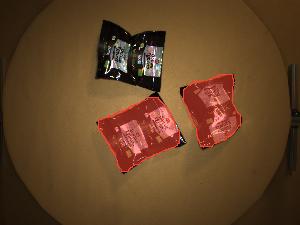} &
    \includegraphics[width=0.14\textwidth]{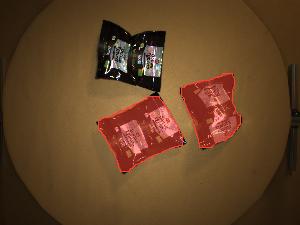} &
    \includegraphics[width=0.14\textwidth]{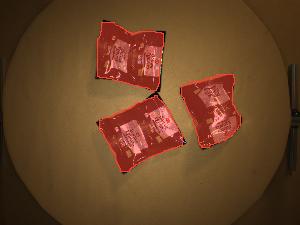} &
    \includegraphics[width=0.14\textwidth]{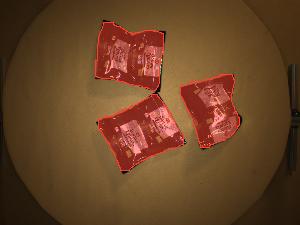} &
    \includegraphics[width=0.14\textwidth]{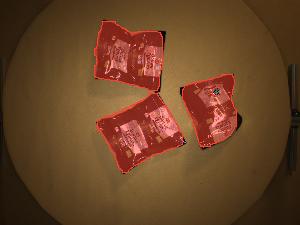} &
    \includegraphics[width=0.14\textwidth]{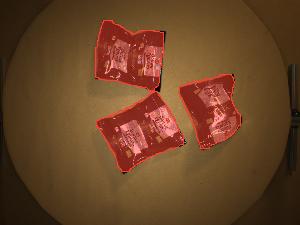}
    \\
    \includegraphics[width=0.14\textwidth]{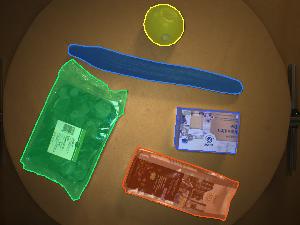} &
    \includegraphics[width=0.14\textwidth]{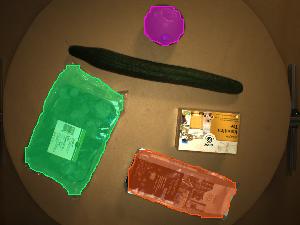} &
    \includegraphics[width=0.14\textwidth]{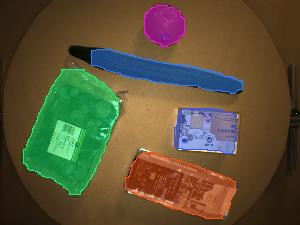} &
    \includegraphics[width=0.14\textwidth]{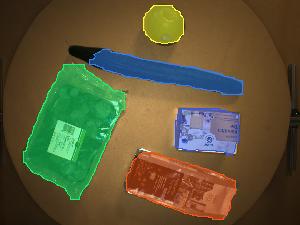} &
    \includegraphics[width=0.14\textwidth]{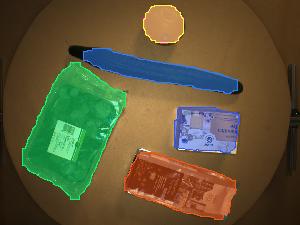} &
    \includegraphics[width=0.14\textwidth]{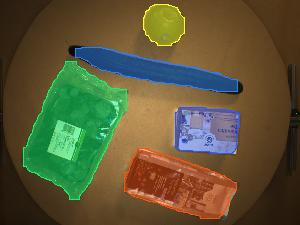} &
    \includegraphics[width=0.14\textwidth]{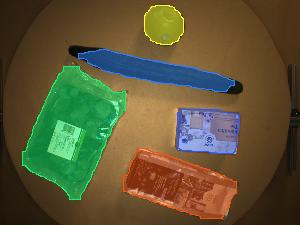}
    \\
    \includegraphics[width=0.14\textwidth]{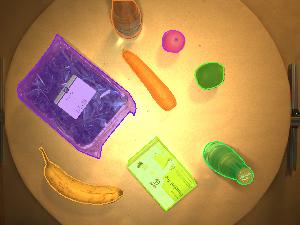} &
    \includegraphics[width=0.14\textwidth]{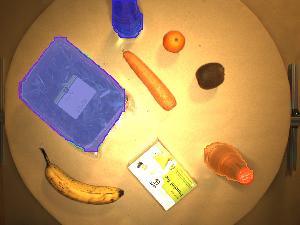} &
    \includegraphics[width=0.14\textwidth]{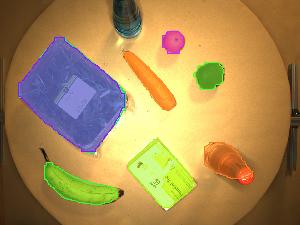} &
    \includegraphics[width=0.14\textwidth]{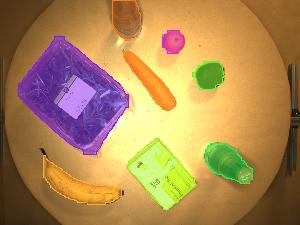} &
    \includegraphics[width=0.14\textwidth]{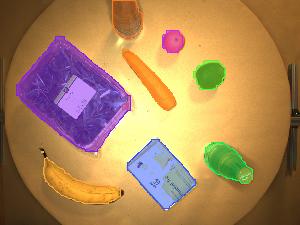} &
    \includegraphics[width=0.14\textwidth]{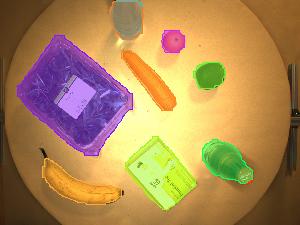} &
    \includegraphics[width=0.14\textwidth]{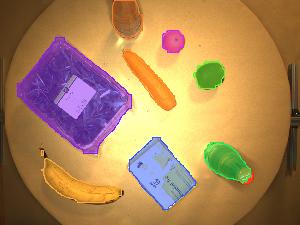}
    \\
    \includegraphics[width=0.14\textwidth]{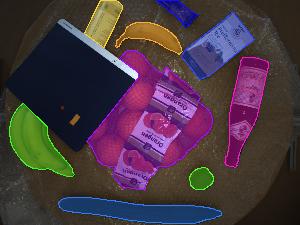} &
    \includegraphics[width=0.14\textwidth]{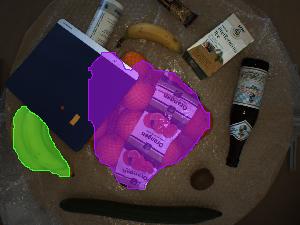} &
    \includegraphics[width=0.14\textwidth]{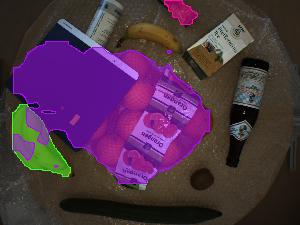} &
    \includegraphics[width=0.14\textwidth]{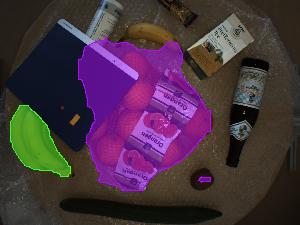} &
    \includegraphics[width=0.14\textwidth]{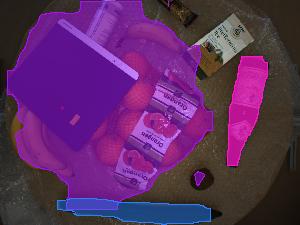} &
    \includegraphics[width=0.14\textwidth]{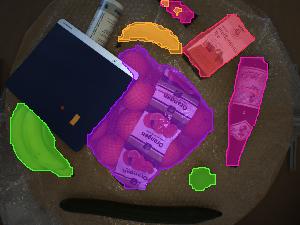} &
    \includegraphics[width=0.14\textwidth]{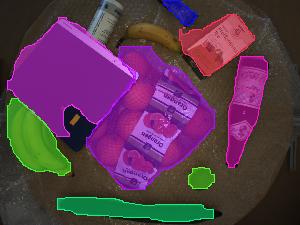}
    \\
    \includegraphics[width=0.14\textwidth]{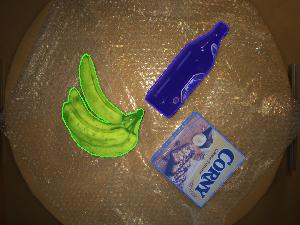} &
    \includegraphics[width=0.14\textwidth]{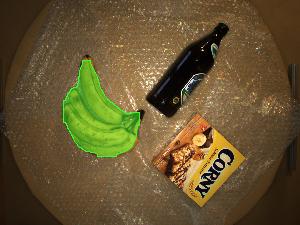} &
    \includegraphics[width=0.14\textwidth]{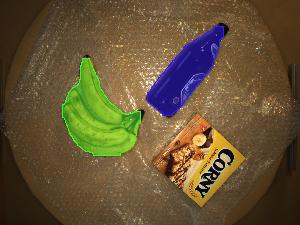} &
    \includegraphics[width=0.14\textwidth]{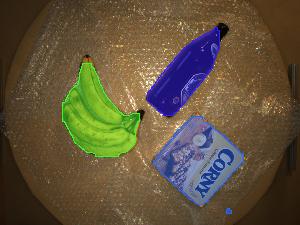} &
    \includegraphics[width=0.14\textwidth]{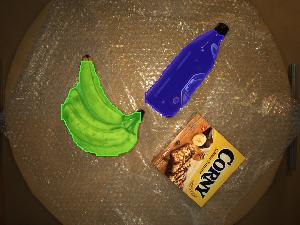} &
    \includegraphics[width=0.14\textwidth]{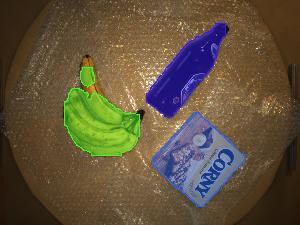} &
    \includegraphics[width=0.14\textwidth]{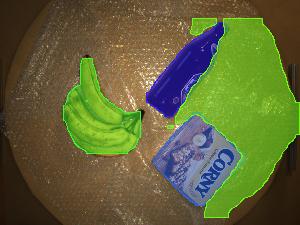}
    \\
    \includegraphics[width=0.14\textwidth]{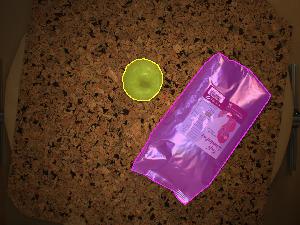} &
    \includegraphics[width=0.14\textwidth]{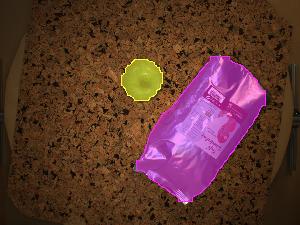} &
    \includegraphics[width=0.14\textwidth]{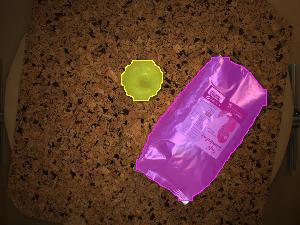} &
    \includegraphics[width=0.14\textwidth]{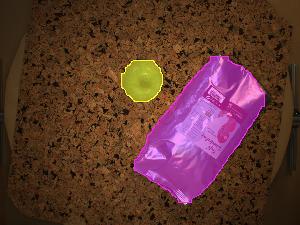} &
    \includegraphics[width=0.14\textwidth]{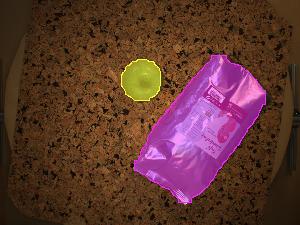} &
    \includegraphics[width=0.14\textwidth]{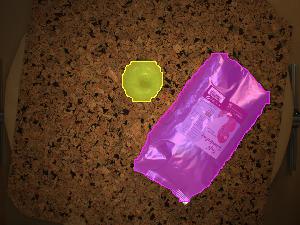} &
    \includegraphics[width=0.14\textwidth]{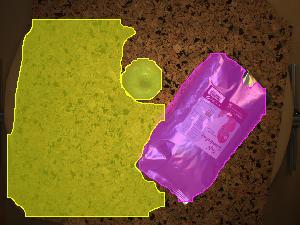}
    \\
    \includegraphics[width=0.14\textwidth]{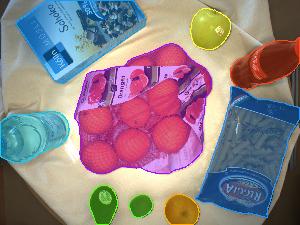} &
    \includegraphics[width=0.14\textwidth]{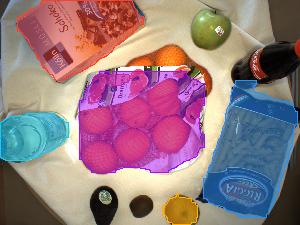} &
    \includegraphics[width=0.14\textwidth]{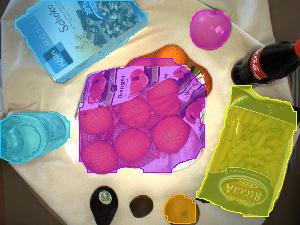} &
    \includegraphics[width=0.14\textwidth]{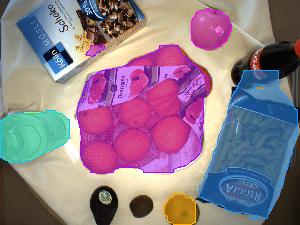} &
    \includegraphics[width=0.14\textwidth]{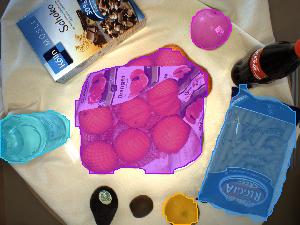} &
    \includegraphics[width=0.14\textwidth]{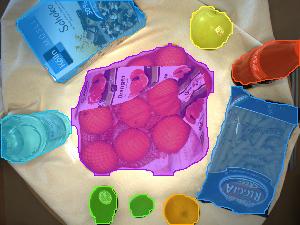} &
    \includegraphics[width=0.14\textwidth]{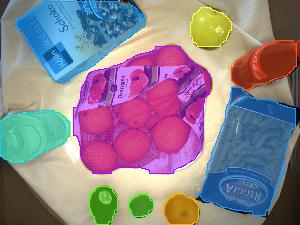}
  \end{tabular}
  \caption{{\bf FCIS qualitative results. }
    A collection of exemplary results of FCIS trained on different splits.
    Note that complex backgrounds often lead to false detections.
    The object classes are color-coded}
  \label{fig:results_fcis}
\end{figure}

\begin{figure}
  \centering
  \scriptsize
  \setlength{\tabcolsep}{0.5pt}
  \begin{tabular}{ccccccc}
    \textbf{ground truth} &
    \texttt{rot0\_light0} &
    \texttt{rot0} &
    \texttt{light0} &
    \texttt{train} &
    \texttt{aug} &
    \texttt{train+aug}
    \\
    \includegraphics[width=0.14\textwidth]{images/qualitative/DSSD_001900_annots.jpg} &
    \includegraphics[width=0.14\textwidth]{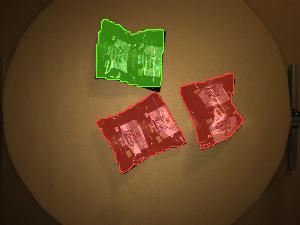} &
    \includegraphics[width=0.14\textwidth]{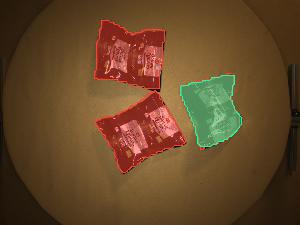} &
    \includegraphics[width=0.14\textwidth]{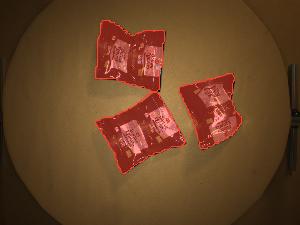} &
    \includegraphics[width=0.14\textwidth]{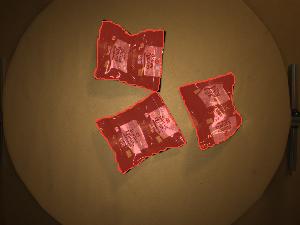} &
    \includegraphics[width=0.14\textwidth]{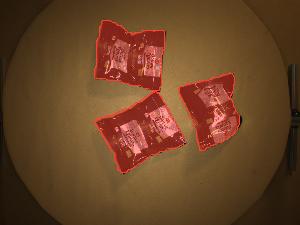} &
    \includegraphics[width=0.14\textwidth]{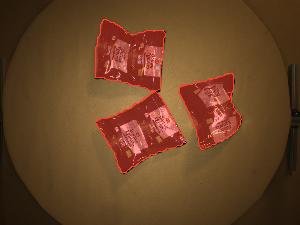}
    \\
    \includegraphics[width=0.14\textwidth]{images/qualitative/DSSD_023100_annots.jpg} &
    \includegraphics[width=0.14\textwidth]{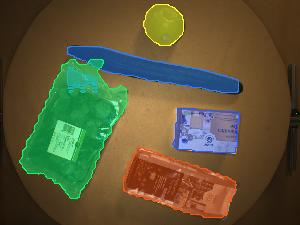} &
    \includegraphics[width=0.14\textwidth]{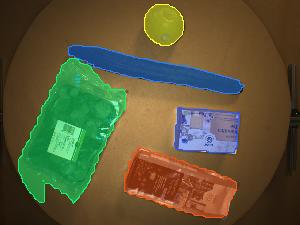} &
    \includegraphics[width=0.14\textwidth]{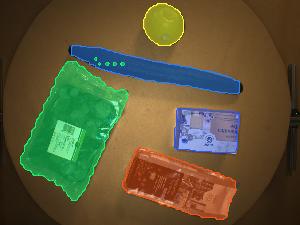} &
    \includegraphics[width=0.14\textwidth]{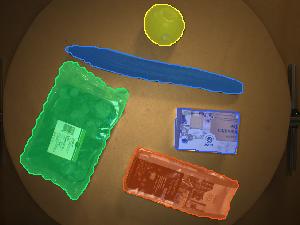} &
    \includegraphics[width=0.14\textwidth]{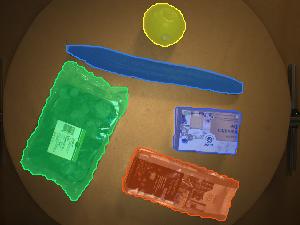} &
    \includegraphics[width=0.14\textwidth]{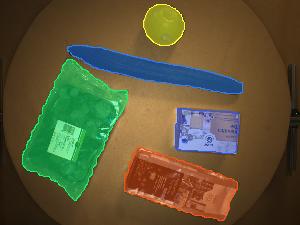}
    \\
    \includegraphics[width=0.14\textwidth]{images/qualitative/DSSD_023510_annots.jpg} &
    \includegraphics[width=0.14\textwidth]{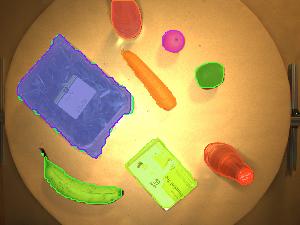} &
    \includegraphics[width=0.14\textwidth]{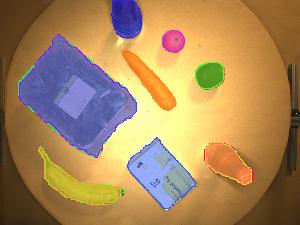} &
    \includegraphics[width=0.14\textwidth]{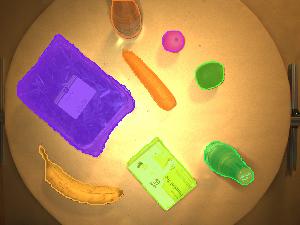} &
    \includegraphics[width=0.14\textwidth]{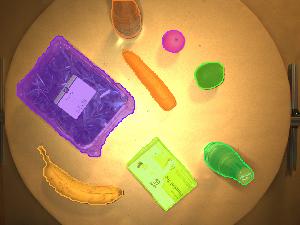} &
    \includegraphics[width=0.14\textwidth]{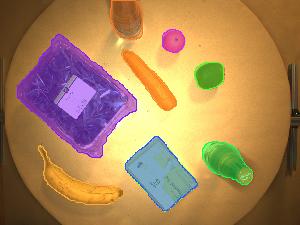} &
    \includegraphics[width=0.14\textwidth]{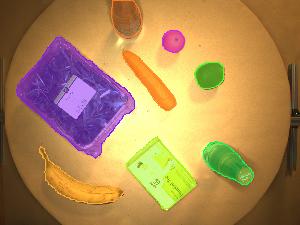}
    \\
    \includegraphics[width=0.14\textwidth]{images/qualitative/DSSD_065920_annots.jpg} &
    \includegraphics[width=0.14\textwidth]{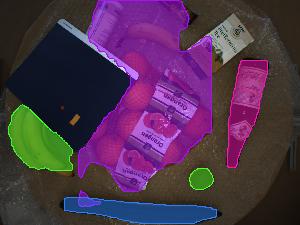} &
    \includegraphics[width=0.14\textwidth]{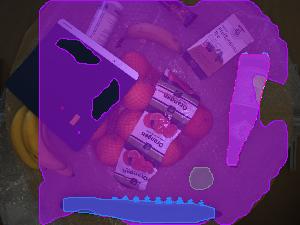} &
    \includegraphics[width=0.14\textwidth]{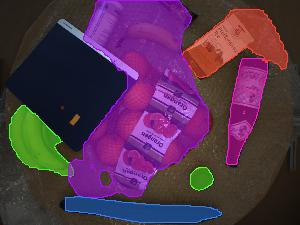} &
    \includegraphics[width=0.14\textwidth]{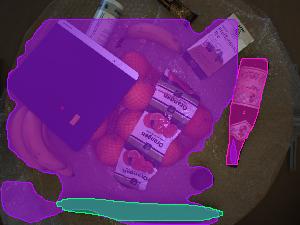} &
    \includegraphics[width=0.14\textwidth]{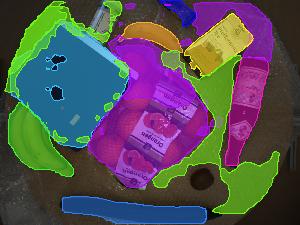} &
    \includegraphics[width=0.14\textwidth]{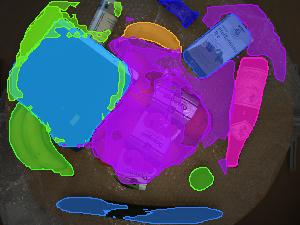}
    \\
    \includegraphics[width=0.14\textwidth]{images/qualitative/DSSD_051400_annots.jpg} &
    \includegraphics[width=0.14\textwidth]{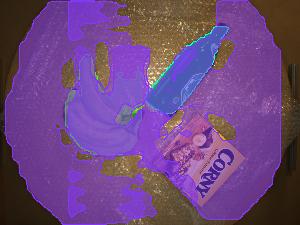} &
    \includegraphics[width=0.14\textwidth]{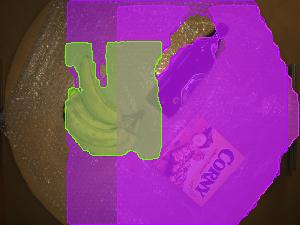} &
    \includegraphics[width=0.14\textwidth]{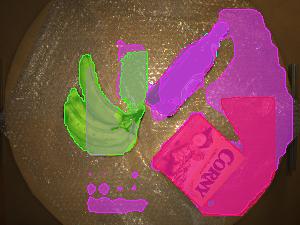} &
    \includegraphics[width=0.14\textwidth]{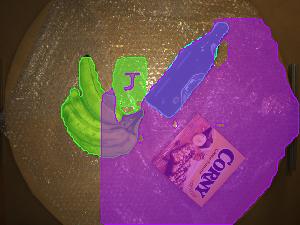} &
    \includegraphics[width=0.14\textwidth]{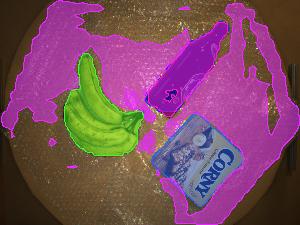} &
    \includegraphics[width=0.14\textwidth]{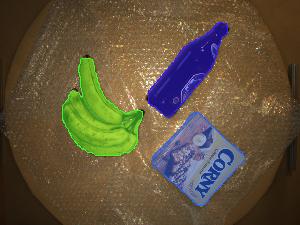}
    \\
    \includegraphics[width=0.14\textwidth]{images/qualitative/DSSD_060600_annots.jpg} &
    \includegraphics[width=0.14\textwidth]{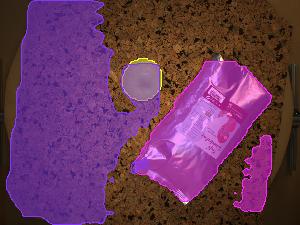} &
    \includegraphics[width=0.14\textwidth]{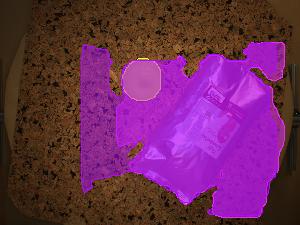} &
    \includegraphics[width=0.14\textwidth]{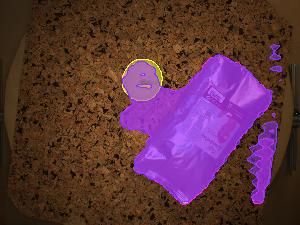} &
    \includegraphics[width=0.14\textwidth]{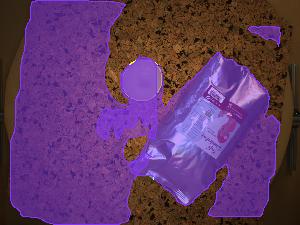} &
    \includegraphics[width=0.14\textwidth]{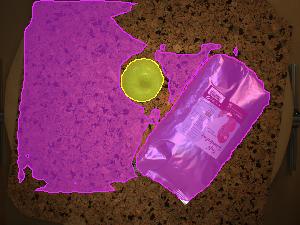} &
    \includegraphics[width=0.14\textwidth]{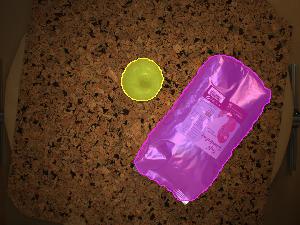}
    \\
    \includegraphics[width=0.14\textwidth]{images/qualitative/DSSD_066900_annots.jpg} &
    \includegraphics[width=0.14\textwidth]{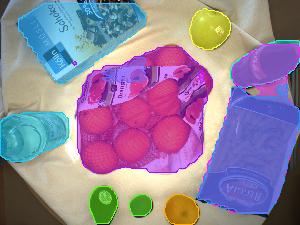} &
    \includegraphics[width=0.14\textwidth]{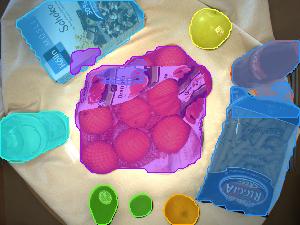} &
    \includegraphics[width=0.14\textwidth]{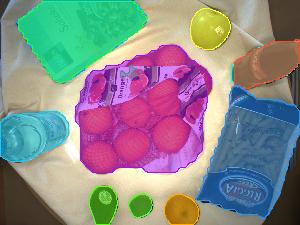} &
    \includegraphics[width=0.14\textwidth]{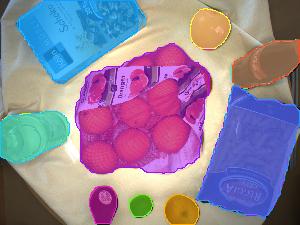} &
    \includegraphics[width=0.14\textwidth]{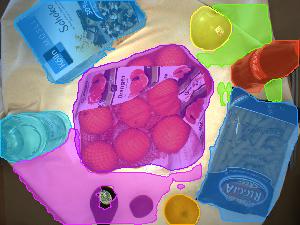} &
    \includegraphics[width=0.14\textwidth]{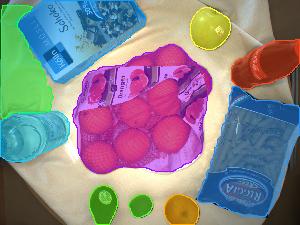}
  \end{tabular}
  \caption{{\bf Mask R-CNN qualitative results. }
    A collection of exemplary results of Mask R-CNN trained on different
    splits. Note that complex backgrounds often lead to false detections.
    The object classes are color-coded}
  \label{fig:results_mrcnn}
\end{figure}

Figures \ref{fig:results_fcis} and \ref{fig:results_mrcnn} show qualitative
instance segmentation results of FCIS and Mask R-CNN, respectively. The models
have been trained on different splits of the D2S dataset. Note that the model
trained on \texttt{train+aug} performs significantly better than the model
trained only on \texttt{train}, especially for objects with occlusion. All
models struggle with textured background and reflections.

\section{Data Augmentation}

The \texttt{aug} split consists of 10\,000 artificially generated images. 
In order to evaluate the influence of the number of training images, we used 
subsets of 2\,000, 4\,000, 6\,000, and 8\,000 images of \texttt{aug}, respectively, 
to train FCIS pretrained on ImageNet. The resulting AP values for both
validation and test set are shown in \figref{fig:dssd_augmentation_influence}.
It can be seen that using more images results in better performance. However,
already with 2\,000 generated images the AP increases by over 20 percentage
points compared to using \texttt{train}.

\section{Per-Class Results}
Tables \ref{tbl:class_aps_fcis} and \ref{tbl:class_aps_mrcnn} show the
per-class AP values for FCIS and Mask R-CNN, respectively, for all 60 classes
and all training splits.  In general, the highest APs are achieved for classes
with large objects and no reflections, such as \textit{lettuce, salad\_iceberg}
or \textit{oranges}.
The data augmentation boosts particularly the worst-performing classes, such as
\textit{banana\_single, cucumber, zucchini} and all
\textit{gepa\_bio\_und\_fair\_$^*$}-classes.

\section{Class Tree}
\begin{figure*}
  \centering
  \includegraphics[width=1.0\textwidth]{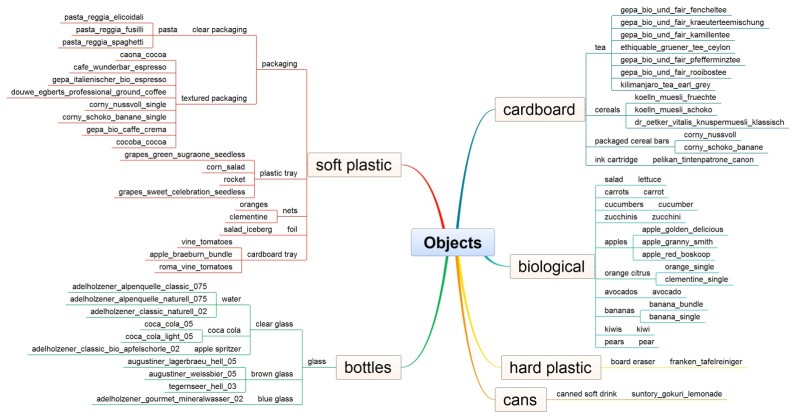} 
  \caption{{\bf Class Tree.}
    Objects are sorted based mainly on their appearance and packaging rather
    than on their content. For example, different kinds of coffee are close to
    each other since they all have a textured, soft plastic packaging. But they
    are far from all sorts of tea, which is packaged in cardboard.}
  \label{fig:ClassTree}
\end{figure*}

The class tree assigns superclasses to classes according to their appearance or
packaging in a hierarchical manner. It is visualized in \figref{fig:ClassTree}.

\begin{table*}
  \caption{{\bf FCIS AP values per class.}
    The AP values of the FCIS models trained on different training sets are
    calculated on the \texttt{test} set. Note that the mean AP is calculated
    over all 60 classes. For each training set, the highest class AP value is
    highlighted in bold.}
  \label{tbl:class_aps_fcis}
  \centering
  \def\arraystretch{1.025} 
  \scriptsize
  \newcommand{\mybox}[1]{\makebox[11mm][c]{\texttt{#1}}}
  \begin{tabular}{lccccccc}
    class & \mybox{train} & \mybox{rot0} & \mybox{light0} &
    \mybox{\scalebox{.85}[1.0]{rot0\_light0}} & \mybox{aug} & \mybox{train+aug}\\ \hline
    adelholzener\_alpenquelle\_classic\_075&40.1&23.9&28.1&17.4&42.9&51.8\\
    adelholzener\_alpenquelle\_naturell\_075&46.7&27.9&36.7&23.8&68.0&69.1\\
    adelholzener\_classic\_bio\_apfelschorle\_02&49.6&24.0&47.9&8.0&64.8&73.7\\
    adelholzener\_classic\_naturell\_02&47.6&18.9&49.3&10.1&62.6&71.1\\
    adelholzener\_gourmet\_mineralwasser\_02&54.0&22.4&55.3&15.6&71.8&78.8\\
    augustiner\_lagerbraeu\_hell\_05&49.9&26.2&47.0&17.1&52.8&60.2\\
    augustiner\_weissbier\_05&21.4&17.1&20.0&7.2&33.6&50.5\\
    coca\_cola\_05&45.1&24.5&42.6&18.2&56.8&57.0\\
    coca\_cola\_light\_05&43.6&11.4&34.2&8.1&55.0&60.8\\
    suntory\_gokuri\_lemonade&52.8&49.6&54.6&44.6&62.1&72.1\\
    tegernseer\_hell\_03&36.9&8.8&37.2&7.7&42.3&64.8\\
    corny\_nussvoll&51.6&53.6&55.8&43.1&89.9&91.6\\
    corny\_nussvoll\_single&59.0&22.9&59.6&16.5&68.0&72.0\\
    corny\_schoko\_banane&44.1&21.4&40.4&17.3&83.2&85.1\\
    corny\_schoko\_banane\_single&17.4&13.7&23.1&11.0&54.0&54.6\\
    dr\_oetker\_vitalis\_knuspermuesli\_klassisch&46.3&21.7&27.5&16.8&80.0&83.1\\
    koelln\_muesli\_fruechte&47.0&29.7&38.5&24.6&82.2&85.5\\
    koelln\_muesli\_schoko&46.2&11.0&30.1&12.1&76.8&80.1\\
    caona\_cocoa&57.6&63.3&62.4&67.9&85.0&85.0\\
    cocoba\_cocoa&41.6&37.6&44.7&39.9&72.3&75.4\\
    cafe\_wunderbar\_espresso&45.5&43.6&53.1&34.5&82.3&74.0\\
    douwe\_egberts\_professional\_ground\_coffee&41.7&37.3&40.6&30.8&71.8&68.7\\
    gepa\_bio\_caffe\_crema&40.7&36.3&43.5&38.6&62.4&64.8\\
    gepa\_italienischer\_bio\_espresso&52.4&32.9&49.9&36.5&59.6&66.7\\
    apple\_braeburn\_bundle&58.9&68.2&66.0&67.9&84.7&88.9\\
    apple\_golden\_delicious&53.6&43.6&57.3&40.8&80.5&85.8\\
    apple\_granny\_smith&58.4&54.8&64.0&52.9&78.8&82.6\\
    apple\_red\_boskoop&63.4&67.3&65.9&51.0&82.1&86.2\\
    avocado&55.6&56.4&54.9&52.5&83.4&84.4\\
    banana\_bundle&62.9&66.0&\textbf{73.1}&62.3&79.0&80.8\\
    banana\_single&23.9&19.8&26.3&16.3&51.6&51.4\\
    clementine&50.8&59.9&58.9&66.2&80.3&81.6\\
    clementine\_single&48.2&50.1&48.6&40.8&68.8&72.8\\
    grapes\_green\_sugraone\_seedless&52.8&61.7&57.4&59.5&76.8&78.3\\
    grapes\_sweet\_celebration\_seedless&55.7&55.5&63.3&58.4&76.8&80.2\\
    kiwi&31.8&27.6&31.5&24.1&72.0&75.4\\
    orange\_single&58.1&71.2&61.4&69.4&76.7&82.8\\
    oranges&62.5&65.8&65.0&66.5&80.8&83.5\\
    pear&60.2&45.2&59.3&27.7&75.7&77.9\\
    pasta\_reggia\_elicoidali&49.0&31.1&53.9&29.5&88.4&86.7\\
    pasta\_reggia\_fusilli&49.4&30.6&54.4&30.9&77.5&53.2\\
    pasta\_reggia\_spaghetti&47.3&36.0&47.5&32.8&82.3&83.4\\
    franken\_tafelreiniger&47.3&37.3&40.7&22.1&48.3&62.7\\
    pelikan\_tintenpatrone\_canon&22.3&18.5&22.4&9.2&66.5&65.6\\
    ethiquable\_gruener\_tee\_ceylon&53.1&44.7&54.0&34.3&70.5&80.2\\
    gepa\_bio\_und\_fair\_fencheltee&14.9&9.4&13.8&5.3&55.7&57.6\\
    gepa\_bio\_und\_fair\_kamillentee&26.5&21.4&27.4&13.3&64.1&50.3\\
    gepa\_bio\_und\_fair\_kraeuterteemischung&29.2&15.3&19.8&13.2&67.4&61.5\\
    gepa\_bio\_und\_fair\_pfefferminztee&36.5&18.2&22.4&11.6&68.8&64.9\\
    gepa\_bio\_und\_fair\_rooibostee&44.0&28.3&44.5&21.4&70.0&75.8\\
    kilimanjaro\_tea\_earl\_grey&40.6&28.7&41.5&26.6&63.7&70.6\\
    cucumber&27.6&18.1&29.3&13.6&60.6&58.6\\
    carrot&36.1&40.7&37.1&36.9&54.6&53.4\\
    corn\_salad&47.1&37.6&50.3&42.6&78.9&79.7\\
    lettuce&\textbf{67.3}&\textbf{79.5}&70.6&\textbf{81.1}&\textbf{91.7}&\textbf{92.2}\\
    vine\_tomatoes&39.0&45.3&49.4&51.7&71.4&70.9\\
    roma\_vine\_tomatoes&41.1&35.3&46.1&27.9&66.7&68.4\\
    rocket&45.4&28.8&52.3&40.3&68.7&70.9\\
    salad\_iceberg&53.9&63.7&61.7&69.5&81.4&83.5\\
    zucchini&44.1&29.2&44.6&24.3&64.4&71.3
  \end{tabular}
\end{table*}

\begin{table*}
  \caption{{\bf Mask R-CNN AP values per class.}
    The AP values of the Mask R-CNN models trained on different training sets
    are calculated on the \texttt{test} set. Note that the mean AP is
    calculated over all 60 classes. For each training set, the highest class
    AP value is highlighted in bold.}
  \label{tbl:class_aps_mrcnn}
  \centering
  \def\arraystretch{1.025} 
  \scriptsize
  \newcommand{\mybox}[1]{\makebox[11mm][c]{\texttt{#1}}}
  \begin{tabular}{lccccccc}
    class & \mybox{train} & \mybox{rot0} & \mybox{light0} &
    \mybox{\scalebox{.85}[1.0]{rot0\_light0}} & \mybox{aug} & \mybox{train+aug}\\ \hline
    adelholzener\_alpenquelle\_classic\_075 &40.6&16.4&28.2&9.4&43.1&44.8\\
    adelholzener\_alpenquelle\_naturell\_075 &53.8&23.8&37.1&23.1&72.1&76.3\\
    adelholzener\_classic\_bio\_apfelschorle\_02 &57.7&12.1&56.7&4.7&65.8&75.9\\
    adelholzener\_classic\_naturell\_02 &48.5&11.3&59.3&36.3&68.4&79.0\\
    adelholzener\_gourmet\_mineralwasser\_02 &69.9&36.9&61.4&45.7&76.2&86.8\\
    augustiner\_lagerbraeu\_hell\_05 &48.6&17.1&57.0&17.9&49.8&70.1\\
    augustiner\_weissbier\_05 &15.6&12.7&21.9&13.0&41.2&50.5\\
    coca\_cola\_05 &45.8&23.7&48.8&16.4&53.2&61.2\\
    coca\_cola\_light\_05 &38.9&16.7&35.1&9.1&55.3&70.8\\
    suntory\_gokuri\_lemonade &67.1&51.6&67.2&54.0&59.4&83.0\\
    tegernseer\_hell\_03 &44.8&11.7&38.5&6.0&47.7&67.8\\
    corny\_nussvoll &53.6&45.1&49.2&55.3&93.0&\textbf{96.6}\\
    corny\_nussvoll\_single &70.9&40.8&64.2&32.9&77.6&84.2\\
    corny\_schoko\_banane &45.2&12.5&37.9&29.8&84.1&90.1\\
    corny\_schoko\_banane\_single &23.6&14.1&24.0&19.2&59.7&64.0\\
    dr\_oetker\_vitalis\_knuspermuesli\_klassisch &48.9&28.7&24.8&24.6&83.4&91.0\\
    koelln\_muesli\_fruechte &51.1&24.3&53.8&26.7&88.8&93.3\\
    koelln\_muesli\_schoko &57.8&4.4&38.6&8.1&85.3&90.2\\
    caona\_cocoa &56.9&45.3&44.9&56.0&82.0&88.0\\
    cocoba\_cocoa &51.0&43.0&44.9&36.4&76.1&86.7\\
    cafe\_wunderbar\_espresso &39.3&12.8&44.5&37.1&83.0&88.2\\
    douwe\_egberts\_professional\_ground\_coffee &45.8&38.2&42.2&38.7&71.5&75.1\\
    gepa\_bio\_caffe\_crema &40.4&31.5&41.5&41.9&81.9&82.1\\
    gepa\_italienischer\_bio\_espresso &59.3&30.9&52.5&19.6&61.9&65.5\\
    apple\_braeburn\_bundle &59.8&56.8&60.7&66.2&82.3&91.7\\
    apple\_golden\_delicious &78.2&\textbf{72.8}&66.4&73.0&76.1&91.3\\
    apple\_granny\_smith &72.5&71.4&65.9&66.4&75.5&87.4\\
    apple\_red\_boskoop &\textbf{79.4}&71.2&\textbf{76.6}&\textbf{79.8}&78.1&94.1\\
    avocado &73.0&60.6&76.2&74.2&84.2&93.4\\
    banana\_bundle &51.4&47.5&55.8&46.9&76.8&86.6\\
    banana\_single &27.4&18.0&28.6&20.6&55.9&57.9\\
    clementine &54.6&57.5&58.4&59.8&79.7&83.0\\
    clementine\_single &59.5&65.0&60.4&59.9&81.2&89.7\\
    grapes\_green\_sugraone\_seedless &44.0&29.4&27.8&34.2&68.9&77.2\\
    grapes\_sweet\_celebration\_seedless &61.5&54.9&51.4&46.8&75.7&80.7\\
    kiwi &45.5&60.0&55.6&54.0&68.9&88.7\\
    orange\_single &69.6&70.5&66.5&72.7&81.5&91.2\\
    oranges &61.1&51.1&61.2&65.8&81.3&86.3\\
    pear &56.4&40.7&74.8&70.5&76.5&85.1\\
    pasta\_reggia\_elicoidali &43.1&23.3&48.2&38.8&88.5&90.5\\
    pasta\_reggia\_fusilli &48.8&23.7&51.6&38.4&76.7&76.5\\
    pasta\_reggia\_spaghetti &55.4&28.3&45.0&28.5&85.0&90.5\\
    franken\_tafelreiniger &48.2&33.4&39.8&34.1&52.8&67.7\\
    pelikan\_tintenpatrone\_canon &30.5&25.6&27.2&32.1&69.1&75.8\\
    ethiquable\_gruener\_tee\_ceylon &52.3&35.9&48.1&18.6&77.8&86.4\\
    gepa\_bio\_und\_fair\_fencheltee &19.3&5.0&14.7&10.9&50.2&55.9\\
    gepa\_bio\_und\_fair\_kamillentee &32.7&17.9&23.3&18.4&64.4&71.1\\
    gepa\_bio\_und\_fair\_kraeuterteemischung &27.0&10.1&17.8&9.4&58.1&72.6\\
    gepa\_bio\_und\_fair\_pfefferminztee &37.5&5.2&24.5&7.9&69.3&77.2\\
    gepa\_bio\_und\_fair\_rooibostee &47.3&29.1&45.0&21.9&70.4&77.3\\
    kilimanjaro\_tea\_earl\_grey &46.1&31.6&38.4&25.4&64.1&75.6\\
    cucumber &29.9&23.9&33.0&27.6&71.4&77.6\\
    carrot &54.1&44.5&42.8&44.5&62.7&72.1\\
    corn\_salad &49.5&33.2&40.5&34.8&77.4&84.0\\
    lettuce &51.6&55.0&34.7&38.4&\textbf{93.4}&96.2\\
    vine\_tomatoes &45.0&39.4&45.2&52.3&71.9&78.0\\
    roma\_vine\_tomatoes &38.8&33.7&37.5&40.2&71.6&73.7\\
    rocket &38.9&15.8&41.0&12.8&71.7&76.1\\
    salad\_iceberg &55.7&53.4&52.9&66.8&77.9&86.4\\
    zucchini &49.4&23.9&51.0&24.8&68.7&84.0
  \end{tabular}
\end{table*}

\begin{figure}
  \centering
  \begin{tikzpicture}
  \begin{axis}[
  scale only axis=true,
  width=4.5cm, height=2cm,
  x label style={at={(axis description cs:0.5,-0.18)},anchor=north},
  y label style={at={(axis description cs:-0.12,.5)},anchor=south},
  xlabel={\# augmented images},
  ylabel={AP},
  xmin=2000, xmax=10000,
  ymin=60, ymax=75,
  scaled x ticks = false,
  xticklabels={0, 2k, 4k, 6k, 8k, 10k},
  legend pos=north west,
  ymajorgrids=true,
  grid style=dashed,
  axis lines=left,
  every axis plot/.append style={ultra thick},
  legend style={column sep=2pt, draw=none, line width=1.0pt}, 
  legend columns=2,
  ]
  \addlegendentry{\texttt{val}}
  \addlegendentry{\texttt{test}}
  \addplot [color=colorAqua,mark=*] table {data/aug_map_val.dat};
  \addplot [color=colorOrange,mark=*] table {data/aug_map_test.dat};
  \end{axis}
  \end{tikzpicture}
  \caption{{\bf Data augmentation.} Influence of number of artificially generated training images on AP for validation and test set}
  \label{fig:dssd_augmentation_influence}
\end{figure}




%
%
%
\bibliographystyle{splncs04}
\bibliography{eccv18_d2s}

\end{document}

%% file: section_intro.tex
\section{Introduction}
The task of \emph{instance-aware semantic segmentation} (\emph{instance
segmentation} for short) can be interpreted as the combination of
\emph{semantic segmentation} and \emph{object detection}.
While \emph{semantic segmentation} methods predict a semantic category for each
pixel \cite{long_2015_fully}, \emph{object detection} focuses on generating
bounding boxes for all object instances within an image \cite{ren_2015_faster}.
As a combination of both, \emph{instance segmentation} methods generate
pixel-precise masks for all object instances in an image.
While solving this task was considered a distant dream a few years ago, the
recent advances in computer vision have made instance segmentation a key focus
of current research \cite{he_2017_mask,li_2016_fully,long_2015_fully}.
This is especially due to the progress in deep convolutional networks
\cite{lecun_2015_deep} and the development of strong baseline frameworks such
as Faster R-CNN \cite{ren_2015_faster} and Fully Convolutional Networks (FCN)
\cite{long_2015_fully}.

\paragraph{Related Work. }All top-performing methods in common instance segmentation challenges are
based on deep learning and require a large amount of annotated training data.
Accordingly, the availability of large-scale datasets, such as
\emph{ADE20K}~\cite{zhou_2016_ade20k},
\emph{Cityscapes}~\cite{cordts_2016_cityscapes},
\emph{ImageNet}~\cite{Russakovsky_2015_imagenet},
\emph{KITTI}~\cite{geiger_2013_kitti},
\emph{COCO}~\cite{lin_2014_coco},
\emph{Mapillary Vistas}~\cite{Neuhold_2017_ICCV},
\emph{VOC}~\cite{everingham_2015_pascal},
\emph{Places}~\cite{zhou_2016_places},
\emph{The Plant Phenotyping Datasets}~\cite{MinerviniPRL2015}, or
\emph{Youtube-8M}~\cite{abu_2016_youtube},
is of paramount importance.

Most of the above datasets focus on everyday photography or urban street
scenes, which makes them of limited use for many industrial applications.
Furthermore, the amount and diversity of labeled training data is usually much
lower in industrial settings. To train a visual warehouse system, for instance,
the user typically only has a handful of images of each product in a fixed
setting. Nevertheless, at runtime, the products need to be robustly detected in
very diverse settings.

With the availability of depth sensors a number of dedicated RGBD datasets have been published \cite{lai2014unsupervised,lai2011large,rennie2016dataset,richtsfeld2012segmentation}: In comparison, these datasets are designed for pose estimation and generally have low resolution images. They often contain fewer scenes (e.g. 111 for \cite{richtsfeld2012segmentation}) that are captured with video \cite{lai2011large} resulting in a high number of frames. Some datasets provide no class annotations \cite{richtsfeld2012segmentation}. \cite{lai2011large} shows fewer, but similar categories to D2S, but only single objects are captured and annotated with lower quality segmentations. In \cite{lai2014unsupervised}, some of these objects occur in real scenes but only box-annotations are provided. The most similar to D2S is \cite{rennie2016dataset}: CAD-models and  object poses are available and could be used to generate ground truth segmentation masks for non-deformable objects. Compared to D2S, the dataset does not show scenes with more than one instance of the same category and objects appear at a much lower resolution.

Only few datasets focus on industry-relevant challenges in the context of
warehouses.
The Freiburg Groceries Dataset \cite{jund2016freiburg},
SOIL-47 \cite{koubaroulis2002evaluating}, and
the Supermarket Produce Dataset \cite{rocha2010automatic}
contain images of supermarket products, but only provide class annotations on
image level, and hence no segmentation.
The Grocery Products Dataset \cite{george2014recognizing} and
GroZi-120 \cite{merler2007recognizing}
include bounding box annotations that can be used for object detection.
However, not all object instances in the images are labeled separately. 
To the best of our knowledge, none of the existing industrial datasets
provides pixel-wise annotations on instance level.
In this paper, we introduce the \emph{Densely Segmented Supermarket (D2S)
dataset}, which satisfies the industrial requirements described above.
The training, validation, and test sets are explicitly designed to
resemble the real-world applications of automatic checkout, inventory, or
warehouse systems.

\paragraph{Contributions.}
We present a novel instance segmentation dataset with high-resolution images in
a real-world, industrial setting. The annotations for the 60 different object
categories were obtained in a meticulous labeling process and are of very high
quality. Specific care was taken to ensure that every occurring instance is
labeled comprehensively.
We show that the high-quality region annotations of the training set can easily
be used for artificial data augmentation.
Using both the original training data and the augmented
data leads to a significant improvement of the average precision (AP) on
the test set by about 30 percentage points.
%
In contrast to existing datasets, our setup and the choice of the objects
ensures that there is no ambiguity in the labels and an AP of 100\% is
achievable by an algorithm that performs flawlessly. To evaluate the
generalizability of methods, the training set is considerably smaller than the
validation and test sets and contains mainly images that show instances of a
single category on a homogeneous background.
Overall, the dataset serves as a demanding benchmark and resembles real-world
applications and their challenges. The dataset is publicly
available\footnote{\url{https://www.mvtec.com/research}}.

%% file: section_dataset.tex
\begin{figure}
  \centering
  \includegraphics[width=0.09\textwidth,height=0.09\textwidth]{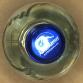}
  \includegraphics[width=0.09\textwidth,height=0.09\textwidth]{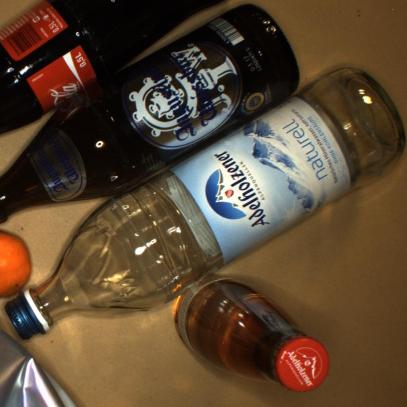}
  \includegraphics[width=0.09\textwidth,height=0.09\textwidth]{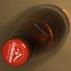}
  \includegraphics[width=0.09\textwidth,height=0.09\textwidth]{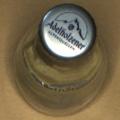}
  \includegraphics[width=0.09\textwidth,height=0.09\textwidth]{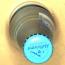}
  \includegraphics[width=0.09\textwidth,height=0.09\textwidth]{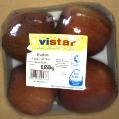}
  \includegraphics[width=0.09\textwidth,height=0.09\textwidth]{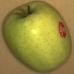}
  \includegraphics[width=0.09\textwidth,height=0.09\textwidth]{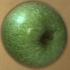}
  \includegraphics[width=0.09\textwidth,height=0.09\textwidth]{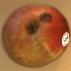}
  \includegraphics[width=0.09\textwidth,height=0.09\textwidth]{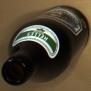}
  \\
  \includegraphics[width=0.09\textwidth,height=0.09\textwidth]{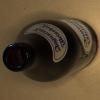}
  \includegraphics[width=0.09\textwidth,height=0.09\textwidth]{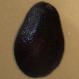}
  \includegraphics[width=0.09\textwidth,height=0.09\textwidth]{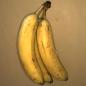}
  \includegraphics[width=0.09\textwidth,height=0.09\textwidth]{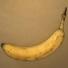}
  \includegraphics[width=0.09\textwidth,height=0.09\textwidth]{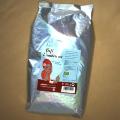}
  \includegraphics[width=0.09\textwidth,height=0.09\textwidth]{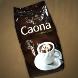}
  \includegraphics[width=0.09\textwidth,height=0.09\textwidth]{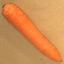}
  \includegraphics[width=0.09\textwidth,height=0.09\textwidth]{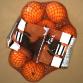}
  \includegraphics[width=0.09\textwidth,height=0.09\textwidth]{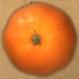}
  \includegraphics[width=0.09\textwidth,height=0.09\textwidth]{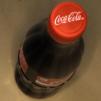}
  \\
  \includegraphics[width=0.09\textwidth,height=0.09\textwidth]{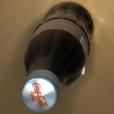}
  \includegraphics[width=0.09\textwidth,height=0.09\textwidth]{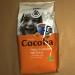}
  \includegraphics[width=0.09\textwidth,height=0.09\textwidth]{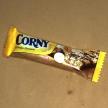}
  \includegraphics[width=0.09\textwidth,height=0.09\textwidth]{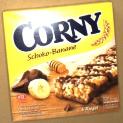}
  \includegraphics[width=0.09\textwidth,height=0.09\textwidth]{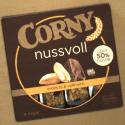}
  \includegraphics[width=0.09\textwidth,height=0.09\textwidth]{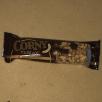}
  \includegraphics[width=0.09\textwidth,height=0.09\textwidth]{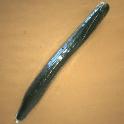}
  \includegraphics[width=0.09\textwidth,height=0.09\textwidth]{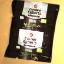}
  \includegraphics[width=0.09\textwidth,height=0.09\textwidth]{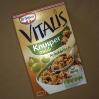}
  \includegraphics[width=0.09\textwidth,height=0.09\textwidth]{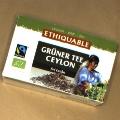}
  \\
  \includegraphics[width=0.09\textwidth,height=0.09\textwidth]{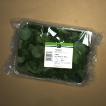}
  \includegraphics[width=0.09\textwidth,height=0.09\textwidth]{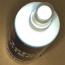}
  \includegraphics[width=0.09\textwidth,height=0.09\textwidth]{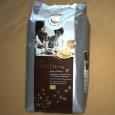}
  \includegraphics[width=0.09\textwidth,height=0.09\textwidth]{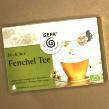}
  \includegraphics[width=0.09\textwidth,height=0.09\textwidth]{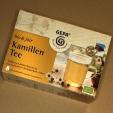}
  \includegraphics[width=0.09\textwidth,height=0.09\textwidth]{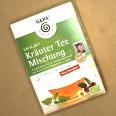}
  \includegraphics[width=0.09\textwidth,height=0.09\textwidth]{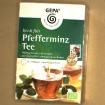}
  \includegraphics[width=0.09\textwidth,height=0.09\textwidth]{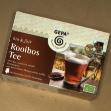}
  \includegraphics[width=0.09\textwidth,height=0.09\textwidth]{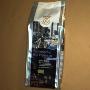}
  \includegraphics[width=0.09\textwidth,height=0.09\textwidth]{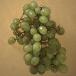}
  \\
  \includegraphics[width=0.09\textwidth,height=0.09\textwidth]{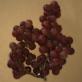}
  \includegraphics[width=0.09\textwidth,height=0.09\textwidth]{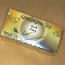}
  \includegraphics[width=0.09\textwidth,height=0.09\textwidth]{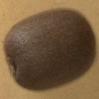}
  \includegraphics[width=0.09\textwidth,height=0.09\textwidth]{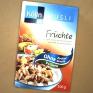}
  \includegraphics[width=0.09\textwidth,height=0.09\textwidth]{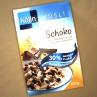}
  \includegraphics[width=0.09\textwidth,height=0.09\textwidth]{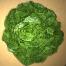}
  \includegraphics[width=0.09\textwidth,height=0.09\textwidth]{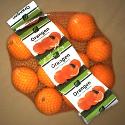}
  \includegraphics[width=0.09\textwidth,height=0.09\textwidth]{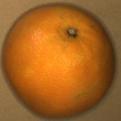}
  \includegraphics[width=0.09\textwidth,height=0.09\textwidth]{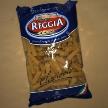}
  \includegraphics[width=0.09\textwidth,height=0.09\textwidth]{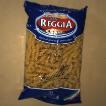}
  \\
  \includegraphics[width=0.09\textwidth,height=0.09\textwidth]{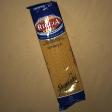}
  \includegraphics[width=0.09\textwidth,height=0.09\textwidth]{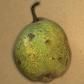}
  \includegraphics[width=0.09\textwidth,height=0.09\textwidth]{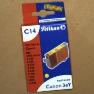}
  \includegraphics[width=0.09\textwidth,height=0.09\textwidth]{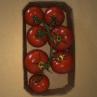}
  \includegraphics[width=0.09\textwidth,height=0.09\textwidth]{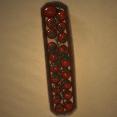}
  \includegraphics[width=0.09\textwidth,height=0.09\textwidth]{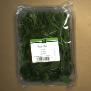}
  \includegraphics[width=0.09\textwidth,height=0.09\textwidth]{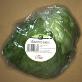}
  \includegraphics[width=0.09\textwidth,height=0.09\textwidth]{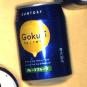}
  \includegraphics[width=0.09\textwidth,height=0.09\textwidth]{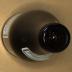}
  \includegraphics[width=0.09\textwidth,height=0.09\textwidth]{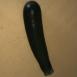}
  \caption{Overview of the 60 different classes within the \emph{D2S} dataset}
  \label{fig:Objects}
\end{figure}

\section{The Densely Segmented Supermarket Dataset}

The overall target of the dataset is to realistically cover the
real-world applications of an automatic checkout, inventory, or
warehouse system. For example, existing automatic checkout systems in
supermarkets identify isolated products that are conveyed on a belt
through a scanning tunnel \cite{ecrs2018,itab2018}. Even though such
systems often provide a semi-controlled environment, external
influences (e.g. lighting changes) cannot be completely
avoided. Furthermore, the system's efficiency is higher if
non-isolated products can be identified as well. Consequently, methods
should be able to segment also partly occluded objects. Also, the
background behind the products is not constant in many applications
because of different types of storage racks in a warehouse system or
because of dirt on the conveyer belt of a checkout system in the
supermarket, for example.

We acquired a total of 21\,000 images in 700 different scenes with various backgrounds, clutter objects, and occlusion levels.
In order to obtain systematic test settings and to reduce the amount of manual
work, a part of the image acquisition process was automated.
Therefore, each scene was rotated ten times with a fixed angle step and
acquired under three different illuminations.

\begin{figure}
  \begin{minipage}{.35\textwidth}
    \centering
    \includegraphics[width=0.74\textwidth]{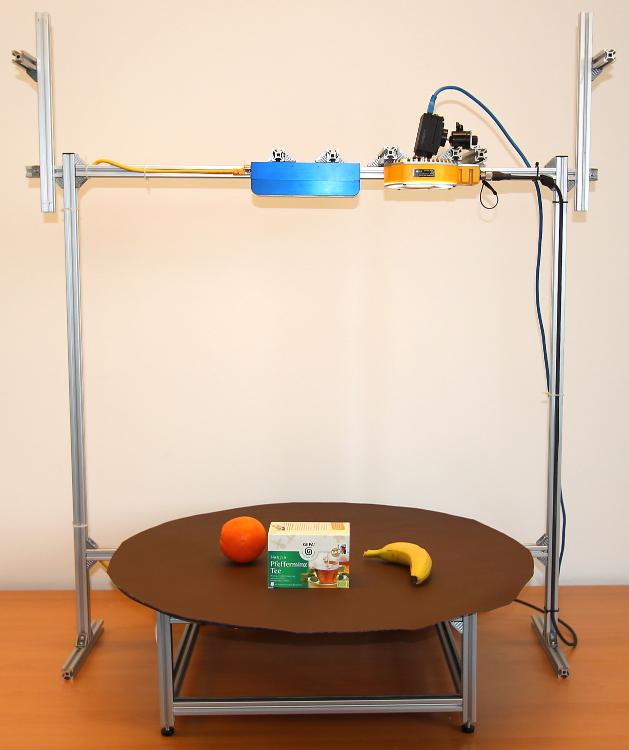}
    \caption{The \emph{D2S} image acquisition setup.
      Each scene was rotated ten times using a turntable.
      For each rotation, three images under different illuminations were
      acquired}
    \label{fig:aufbau}
  \end{minipage}
  \hspace*{\fill}
  \begin{minipage}{.63\textwidth}
    \centering
    \includegraphics[width=0.32\textwidth]{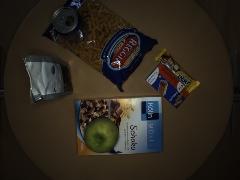}
    \includegraphics[width=0.32\textwidth]{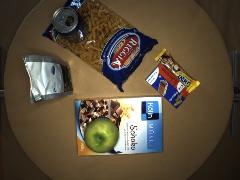}
    \includegraphics[width=0.32\textwidth]{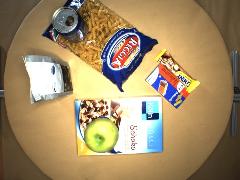} \\
    \vspace{0.1cm}
    \includegraphics[width=0.32\textwidth]{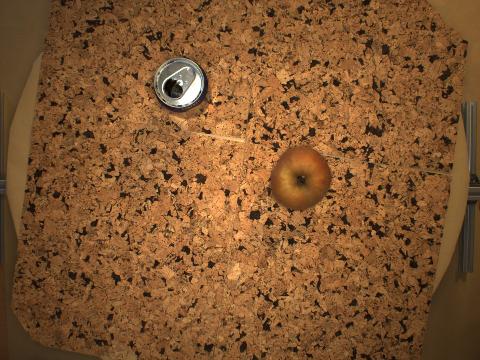}
    \includegraphics[width=0.32\textwidth]{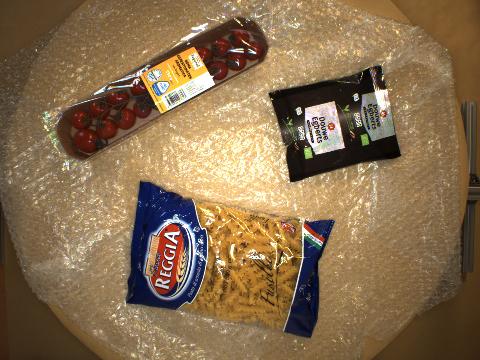}
    \includegraphics[width=0.32\textwidth]{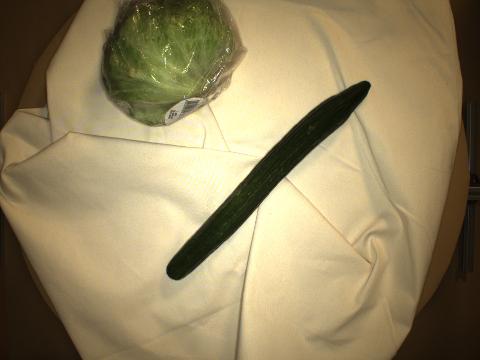}
    \caption{(\emph{Top}) Each scene was acquired under three different lightings.
      (\emph{Bottom}) As opposed to the training set (where a single
      uniform background is used), the test and validation
      sets include three additional backgrounds. This
      allows for a detailed evaluation of the robustness of
      the methods}
    \label{fig:lightings_background}
  \end{minipage}
\end{figure}

\paragraph{Setup.}
The image acquisition setup is depicted in \figref{fig:aufbau}.
A high-resolution ($1920 \times 1440$) industrial color camera was
mounted above a turntable.
The camera was intentionally mounted off-centered with respect to the rotation
center of the turntable to introduce more perspective variations in the rotated images.

\paragraph{Objects.}
An overview of the 60 different classes is shown in \figref{fig:Objects}. The
object categories cover a selection of common, everyday products such as
fruits, vegetables, cereal packets, pasta, and bottles. They are embedded into
a class hierarchy tree that splits the classes into groups
of different packaging. This results in neighboring leafs being visually very
similar, while distant nodes are visually more different, even if they are
semantically similar products, e.g. single apples in comparison to a bundle of
apples in a cardboard tray.
The class hierarchy can be used, for instance, for advanced training and
evaluation strategies similar to those used in \cite{redmon2017yolo9000}. However, it is not used in the scope of
this paper.

\begin{figure}[t]
  \centering
  \includegraphics[width=0.19\textwidth]{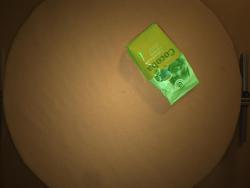}
  \includegraphics[width=0.19\textwidth]{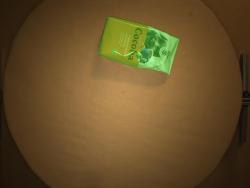}
  \includegraphics[width=0.19\textwidth]{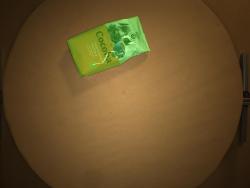}
  \includegraphics[width=0.19\textwidth]{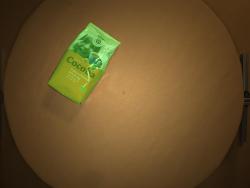}
  \includegraphics[width=0.19\textwidth]{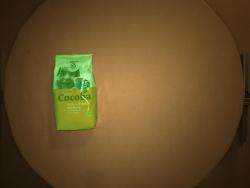}\\
  \vspace*{2pt}
  \includegraphics[width=0.19\textwidth]{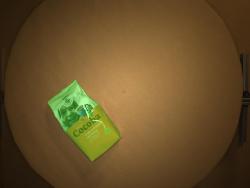}
  \includegraphics[width=0.19\textwidth]{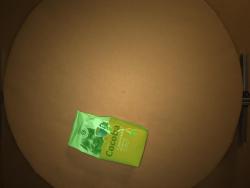}
  \includegraphics[width=0.19\textwidth]{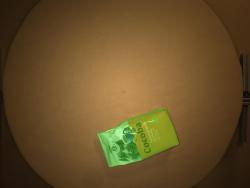}
  \includegraphics[width=0.19\textwidth]{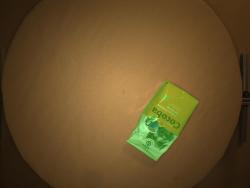}
  \includegraphics[width=0.19\textwidth]{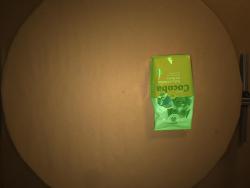}
  \caption{Each scene was acquired at ten different rotations in steps of
           $36^\circ$. The camera was mounted slightly off-centered in order to
           introduce more variation in the images}
  \label{fig:rotations}
\end{figure}

\paragraph{Rotations.}
To increase the number of different views and to evaluate the invariance of approaches with respect to rotations  \cite{follmann_2018_rotational,zhou_2018_orn},
each scene was rotated ten times in increments of $36^\circ$. The turntable
allowed automation and ensured precise rotation angles. An example of the ten rotations for a scene from the training set is
displayed in \figref{fig:rotations}.

\paragraph{Lighting.}
To evaluate the robustness of methods to illumination changes and different
amounts of reflection, each scene and rotation was acquired under three
different lighting settings. For this purpose an LED ring light was attached to
the camera. The illumination was set to span a large spectrum of possible
lightings, from under- to overexposure (see \emph{top} of \figref{fig:lightings_background}).

\paragraph{Background.}
The validation and test scenes have a variety of different backgrounds that are
shown in \figref{fig:lightings_background} (\emph{bottom}). This allows to evaluate
the generalizability of approaches. In contrast, the training set is restricted
to images with a single homogeneous background. It is kept constant to imitate
the settings of a warehouse system, where new products are mostly imaged within
a fixed environment and not in the test scenario.

\begin{figure}
  \begin{minipage}{.48\textwidth}
    \centering
    \includegraphics[width=0.49\textwidth]{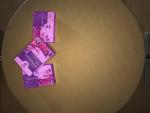}
    \includegraphics[width=0.49\textwidth]{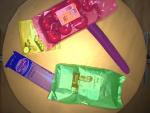}
    \caption{Objects appear with different amounts of occlusion.
      These may either be caused by objects of the same class, objects of a
      different class or by clutter objects not within the training set}
    \label{fig:occlusion}
  \end{minipage}
  \hspace*{\fill}
  \begin{minipage}{.48\textwidth}
    \centering
    \includegraphics[width=0.49\textwidth]{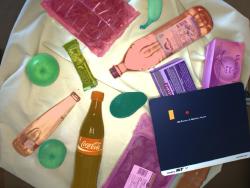}
    \includegraphics[width=0.49\textwidth]{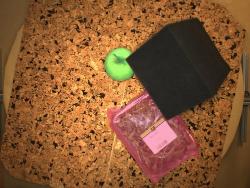}
    \caption{To test the robustness of approaches to unseen clutter objects,
      objects not within the training set were added to the validation and
      test sets (e.g., a mouse pad and a black foam block)}
    \label{fig:clutter}
  \end{minipage}
\end{figure}

\paragraph{Occlusion and Clutter.}
As indicated in \figref{fig:occlusion}, occlusions may arise from objects of the same class, objects of a different class, or from clutter
objects. Clutter objects have a category that is not present in the training
images. They were added explicitly to the validation and test images to
evaluate the robustness to novel objects. Examples of the selected
clutter objects are shown in \figref{fig:clutter}.

%% file: section_splits.tex
\section{Dataset Splitting}

In contrast to existing datasets for instance-aware semantic segmentation,
such as \mbox{\emph{VOC} \cite{everingham_2015_pascal}} and \mbox{\emph{COCO}
\cite{lin_2014_coco}}, 
the \emph{D2S} training set has a different distribution with respect to image
and class statistics than the validation and test sets.
The complexity of the captured scenes as well as the average number of objects
per image are substantially higher in the validation and test sets (see 
\tabref{tbl:split_stats}).
The motivation for this choice of split is to follow common industrial
requirements, such as: low labelling effort, low complexity of training set
acquisition for easy replicability, and the possibility to easily add new
classes to the system. 

The split is performed on a per-scene basis: all 30 images of a scene, i.e. all
combinations of the ten rotations and three lightings, are included in either
the training, the validation, or the test set.
In the following, we describe the rules for generating the splits.

\paragraph{Training Split.}
To meet the mentioned industrial requirements, the training scenes are
selected to be as simple as possible: They have a homogeneous background, mostly contain only one object and the amount of occlusions is reduced to a minimum.
To summarize, we add scenes to the training split that
\begin{itemize} \itemsep0em
  \item contain only objects of one category%
       \footnote{In order to provide similar views of each object class as they
                 are visible in the validation and test set, four scenes were
                 added to the training set that contain two distinct classes.},
  \item provide new views of an object,
  \item only contain objects with no or marginal overlap,
  \item have no clutter and a homogeneous background.
\end{itemize}
%
The total number of scenes in the training set is 147, resulting in 4380 images
of 6900 objects. The rather small training set should encourage work towards
the generation of augmented or synthetic training data, for instance using
generative adversarial networks \cite{Gurumurthy_2017_CVPR,Huang_2017_CVPR,li2017perceptual,shrivastava_2016_learning,Zhang_2017_ICCV}.

\paragraph{Validation and Test Splits.}
The remaining scenes are split between the validation and the test set.
They consist of scenes with
\begin{itemize} \itemsep0em
  \item single or multiple objects of different classes,
  \item touching or occluded objects,
  \item clutter objects and
  \item varying background.
\end{itemize}
These scenes were chosen such that the generalization capabilities of
approaches can be evaluated.
Additionally, current methods struggle with heavy occlusion and novelty
detection.
These issues are addressed by this choice of splits as well.
The split between validation and test set was performed on subgroups of images
containing the same number of total and occluded objects.
This ensures that both sets have approximately the same distribution.
The ratio of the number of scenes in the validation and test set is chosen to
be 1:4. The reasons for this decision are twofold:
First, the evaluation of the model on a small validation set is faster.
Second, we do not want to encourage training on the validation set, but 
stimulate work on approaches that require little training data or use
augmentation techniques.
The statistics of the number of images and objects in the splits are visualized
in \tabref{tbl:split_stats}.

\begin{table}[t]
  \caption{{\bf Split statistics.}
    Due to our splitting strategy, the number of images and the number of
    instances per image is significantly lower for the training set.
    The complexity of validation and test scenes is approximately the same}
  \label{tbl:split_stats}
  \centering
  \def\arraystretch{1.2}
  \setlength{\tabcolsep}{8pt}
  \newcommand{\yes}{\checkmark}
  \begin{tabular}{l r r r r} \\ \hline
    split                  &   \texttt{all} & \texttt{train} &   \texttt{val} &
    \texttt{test} \\ \hline\hline
    \# scenes              &   700 &   146 &   120 &   434 \\
    \# images              & 21000 &  4380 &  3600 & 13020 \\
    \# objects             & 72447 &  6900 & 15654 & 49893 \\
    \# objects/image       &  3.45 &  1.58 &  4.35 &  3.83 \\
    \# scenes w. occlusion &   393 &    10 &    84 &   299 \\
    \# scenes w. clutter   &    86 &     0 &    18 &    68 \\ \hline
    rotations              &       &  \yes &  \yes &  \yes \\
    lighting variation     &       &  \yes &  \yes &  \yes \\
    background variation   &       &       &  \yes &  \yes \\
    clutter                &       &       &  \yes &  \yes \\ \hline
  \end{tabular}
\end{table}

%% file: section_statistics.tex
\section{Statistics \& Comparison}

In this section, we compare our dataset to \emph{VOC} \cite{everingham_2015_pascal} and \emph{COCO} \cite{lin_2014_coco}.
These datasets have encouraged many researchers to work on instance segmentation and are frequently used to benchmark state-of-the-art methods.

\begin{table}[t]
  \caption{{\bf Dataset statistics.}
    Number of images and objects per split, average number of objects per
    image and number of classes for \emph{D2S} (ours), \emph{VOC 2012}, and
    \emph{COCO}. *For \emph{VOC 2012} and \emph{COCO}, the object numbers are
    only available for the training and validation set}
  \label{tbl:dataset_stats}
  \centering
  \def\arraystretch{1.2}
  \setlength{\tabcolsep}{8pt}
  \begin{tabular}{l l r r r} \hline
    \multicolumn{2}{l}{Dataset} & \emph{VOC} & \emph{COCO} & \emph{D2S} \\ \hline\hline
    \# images  & all            &  4369 & 163957 & 21000 \\
               & \texttt{train} &  1464 & 118287 &  4380 \\
               & \texttt{val}   &  1449 &   5000 &  3600 \\
               & \texttt{test}  &  1456 &  40670 & 13020 \\ \hline
    \# objects & all            &  -    & -      & 72447 \\
               & \texttt{train} &  3507 & 849941 &  6900 \\
               & \texttt{val}   &  3422 &  36335 & 15654 \\
               & \texttt{test}  &  -    & -      & 49893 \\ \hline
    \# obj/img &                & 2.38* &  7.19* &  3.45 \\ \hline
    \# classes &                &    20 &     80 &    60 \\ \hline
    \hfill
  \end{tabular}
\end{table}

\paragraph{Dataset Statistics.}
As summarized in \tabref{tbl:dataset_stats}, \emph{D2S} contains significantly
more object instances than \emph{VOC}, but fewer than \emph{COCO}.
Specifically, although the \emph{D2S} training set is larger than that of
\emph{VOC}, the number of training objects is less than 1\% of those in
\emph{COCO}. This choice was made intentionally, as in many industrial
applications it is desired to use as few training images as possible.
In contrast, the proportion of validation images is significantly larger for
\emph{D2S} in order to enable a thorough evaluation of the generalization
capabilities. On average, there are half as many objects per image in
\emph{D2S} as in \emph{COCO}.


\begin{figure}[b]
  \centering
  \includegraphics[width=\textwidth]{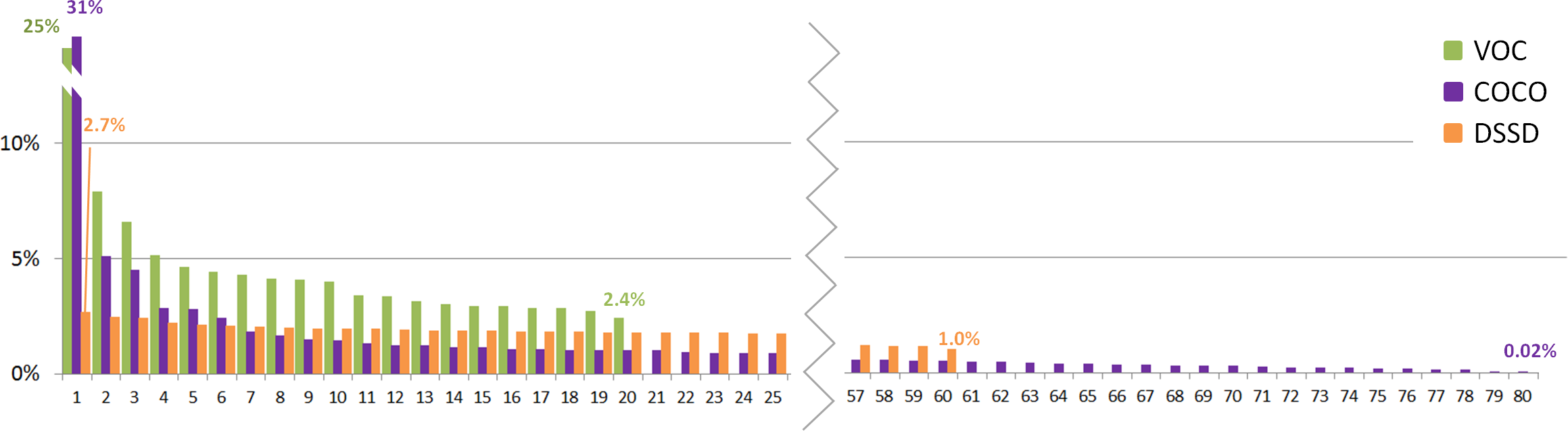}
  \caption{Ratio of objects per class for \emph{D2S} (\emph{orange}), \emph{VOC}
           (\emph{green}) and \emph{COCO} (\emph{violet}). In \emph{COCO} and \emph{VOC},
           the class \emph{person} is dominant and some classes
           are underrepresented. In \emph{D2S}, the number of objects per
           class is uniformly distributed. Note that for \emph{COCO}
           and \emph{VOC} the diagram was calculated based on train and val
           splits}
  \label{fig:class_stats_num_objs}
\end{figure}

\paragraph{Class Statistics.}
Since the images of \emph{COCO} and \emph{VOC} were taken from flickr\footnote{\url{https://www.flickr.com}}, the distribution of object classes is not uniform.
In both datasets, the class \emph{person} dominates, as visualized in
\figref{fig:class_stats_num_objs}:
31\% and 25\% of all objects belong to this class for \emph{COCO} and \emph{VOC}, respectively.
Moreover, 10\% of the classes with the highest number of objects are represented by 51\% and 33\% of all objects, while only 5.4\% and 13.5\% of the objects belong to the 25\% of classes with the lowest number of objects. 
This class imbalance is valid since both \emph{COCO} and \emph{VOC} represent the real world where some classes naturally appear more often than others.
In the evaluation all classes are weighted uniformly.
Therefore, the class imbalance inherently poses a challenge to learn all classes equally well, independent from the number of training samples.
For example, the \emph{COCO 2017} validation set contains nine instances of the
class \emph{toaster}, but 10\,777 instances of \emph{person}.
Nevertheless, both categories are equally weighted in the calculation of the
mean average precision, which is the metric used for ranking the methods in the
\emph{COCO} segmentation challenge. 
 
\begin{figure}[t]
  \centering
  \includegraphics[width=0.95\textwidth]{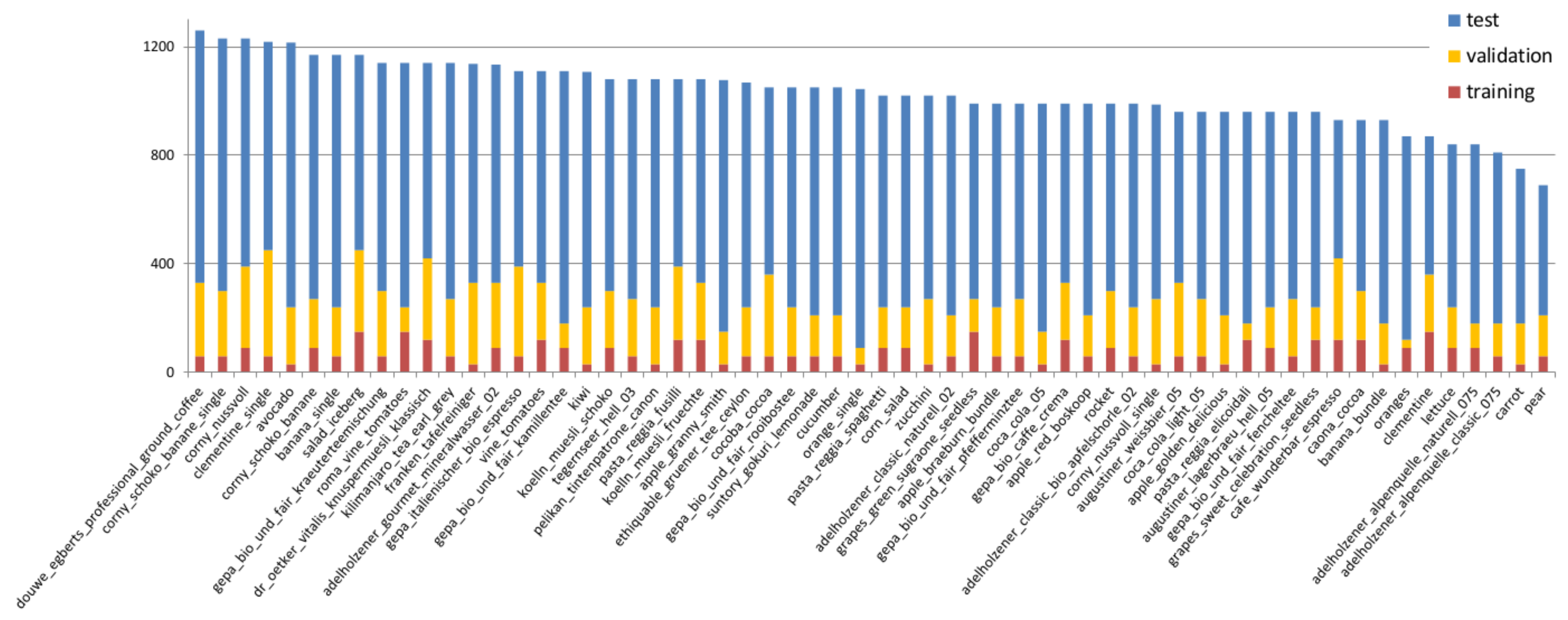}
  \caption{Number of images per class and split sorted by the total number of
    images per class for \emph{D2S}. The number of images per class is almost
    uniformly distributed}
  \label{fig:class_stats_num_imgs_dssd}
\end{figure}

There is no such class imbalance in \emph{D2S}. In the controlled environment
of the supermarket scenario, all classes have the same probability to appear in
an image.
The class with the highest number of objects is represented by only 2.7\% of
all objects. Only 14\% of the objects represent the 10\% of classes with the
highest number of objects, while 19\% of the objects are from the 25\% of
classes with the lowest number of objects.
The class distribution of \emph{D2S} is visualized in
\figref{fig:class_stats_num_imgs_dssd}, where the number of images per class is
shown in total and for each split. 
As mentioned above, the number of images for each class is rather low in the training split, especially for classes that have a similar appearance for different views, such as \emph{kiwi} and \emph{orange\_single}.
Note that, although the split choice between validation and test set is not made on the class level, each class is well represented in both sets. 
The key challenge of the \emph{D2S} dataset is thus not the handling of
underrepresented classes, but the low amount of training data.

\begin{figure}[t]
	\centering
	\includegraphics[width=0.11\textwidth,height=0.11\textwidth]{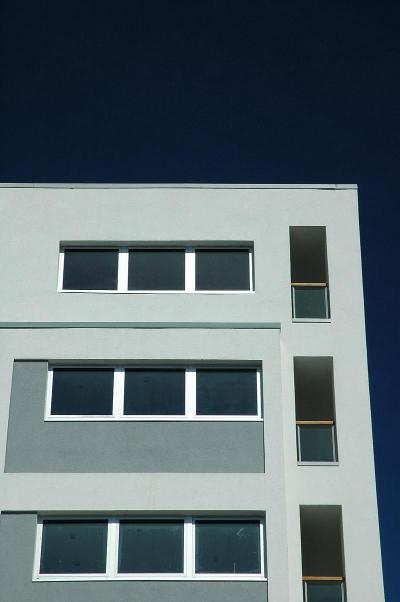}
	\includegraphics[width=0.11\textwidth,height=0.11\textwidth]{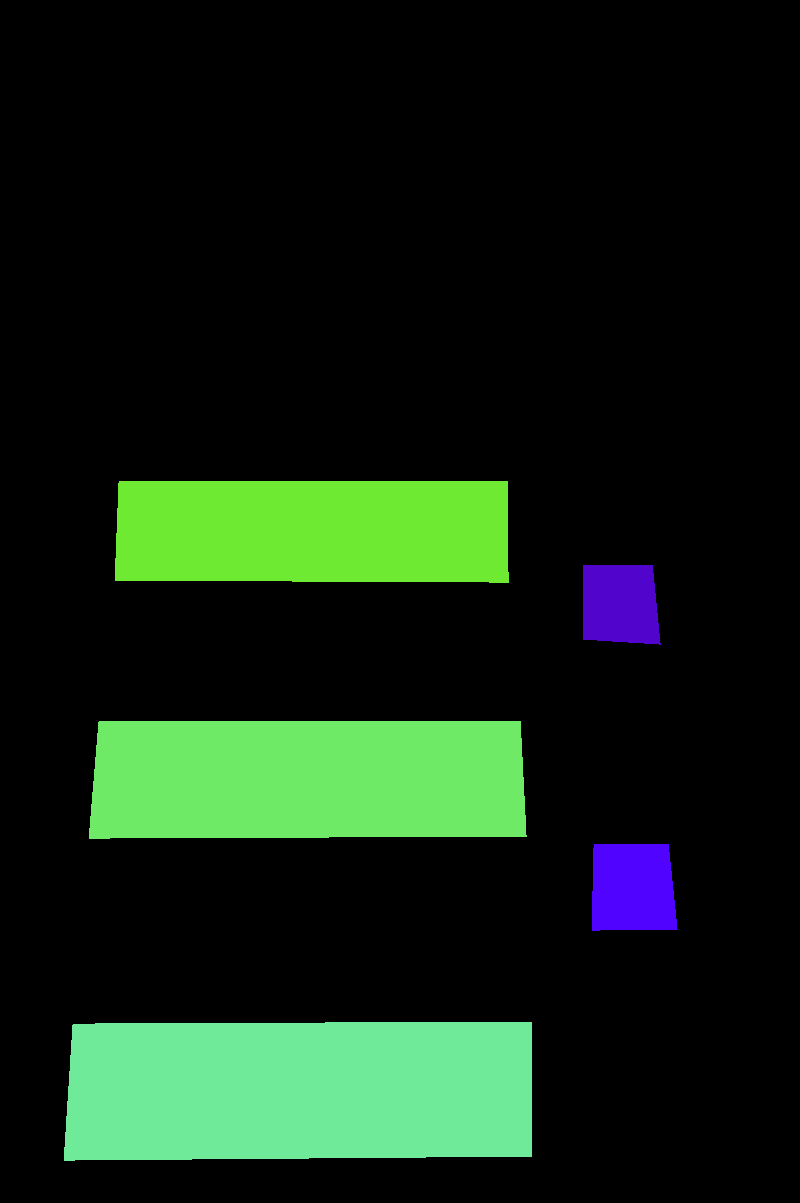}
	\,
	\includegraphics[width=0.11\textwidth,height=0.11\textwidth]{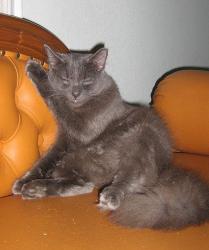}
	\includegraphics[width=0.11\textwidth,height=0.11\textwidth]{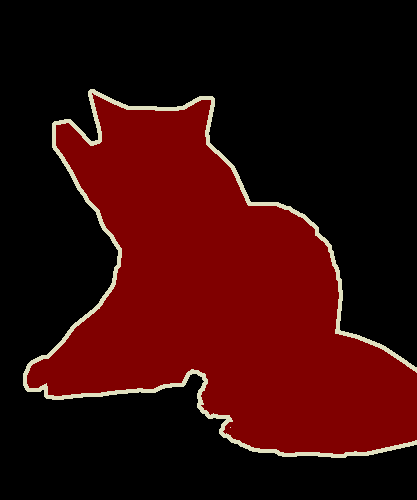}
	\,
	\includegraphics[width=0.11\textwidth,height=0.11\textwidth]{images/labelfailures/coco_donut_missing.jpg}
	\includegraphics[width=0.11\textwidth,height=0.11\textwidth]{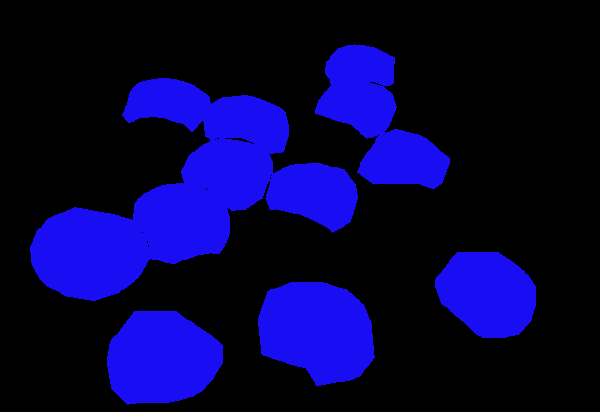}
	\,
	\includegraphics[width=0.11\textwidth,height=0.11\textwidth]{images/labelfailures/coco_person_missing.jpg}
	\includegraphics[width=0.11\textwidth,height=0.11\textwidth]{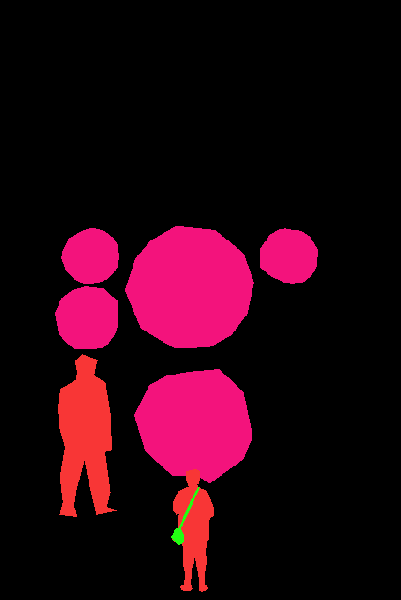}
	\caption{Large real-world datasets are extremely difficult to label
		consistently. In the examples from \emph{ADE20K}, \emph{VOC} and
		\emph{COCO}, some labels are missing (\emph{from left to right}): a window,
		the sofa, some donuts, and the painting of a person}
	\label{fig:LabelFailures}
\end{figure}

\paragraph{Label Consistency.}
It is difficult to ensure that all object instances in large real-world
datasets are labeled consistently.
On the one hand, it is hard to establish a reliable review process for the
labeling of large datasets, e.g. to avoid unlabeled objects.
On the other hand, some labels are ambiguous by nature, for instance a painting of a person.
\figref{fig:LabelFailures} shows examples for label inconsistencies from
\emph{ADE20K} \cite{zhou_2016_ade20k}, \emph{VOC} and \emph{COCO}.

In \emph{D2S}, the object classes are unambiguous and have been labeled by six
expert annotators. All present objects are annotated with high quality labels.
A perfect algorithm, which flawlessly detects and segments every object in all
images of the \emph{D2S} dataset, will achieve an AP of 100\%. This is not the
case for \emph{COCO}, \emph{VOC}, and \emph{ADE20K}. In these datasets, if an
algorithm correctly detects one of the objects that is not labeled, the missing
ground truth leads to a false positive. Furthermore, if such an object is not
found by an algorithm, the resulting false negative is not accounted for. As
algorithms improve, this might prevent better algorithms from obtaining higher
scores in the benchmarks.
In \emph{COCO}, this problem is addressed using
\emph{crowd annotations}, i.e. regions containing many objects of the same
class that are ignored in the evaluation. However, crowd annotations are not
present in all cases.

%% file: section_benchmark.tex
\section{Benchmark}
In this section, we provide first benchmark results for our dataset.
We evaluate the performance of state-of-the-art methods for object detection
\cite{lin2017focal,ren_2015_faster} and instance segmentation
\cite{he_2017_mask,li_2016_fully}.
We experiment with various training sets, which differ in the number of
rotations and the availability of under- and overexposed images.
Furthermore, we evaluate a simple approach for augmenting the training data
artificially.

\subsection{Evaluated Methods}

\paragraph{Object Detection.}
For the object detection task, we evaluate the performance of Faster R-CNN
\cite{ren_2015_faster} and RetinaNet \cite{lin2017focal}.
We use the official implementations of both methods, which are provided in the
Detectron\footnote{\url{https://github.com/facebookresearch/Detectron}}
framework.
Both methods use a ResNet-101 \cite{He_2016_CVPR} backbone with Feature Pyramid
Network \cite{lin2017fpn}.

\paragraph{Instance Segmentation.}
For the instance segmentation task, we evaluate the performance of Mask R-CNN
\cite{he_2017_mask} and FCIS \cite{li_2016_fully}.
We use the official implementation of Mask R-CNN in the Detectron framework and
the official implementation of FCIS provided by the
authors\footnote{\url{https://github.com/msracver/FCIS}}.
Mask R-CNN uses a ResNet-101 with Feature Pyramid Network as backbone, while
FCIS uses a plain ResNet-101.
Since both methods output boxes in addition to the segmentation masks, we also
include them in the object detection evaluation.

\paragraph{Training.}
All methods are trained end-to-end.
The network weights are initialized with the COCO-pretrained models provided
by the respective authors.
The input images are resized to have a shorter side of 800 pixels (600 pixels
for FCIS, respectively).
All methods use horizonal flipping of the images at training time.
FCIS uses \emph{online hard example mining} \cite{shrivastava2016training}
during training.

\subsection{Evaluation Metric} \label{sec:evaluation_metric}
The standard metric used for object detection and instance segmentation is
\emph{mean average precision} (mAP) \cite{everingham_2015_pascal}.
It is used, for instance, for the ranking of state-of-the-art methods in the
\emph{COCO} segmentation challenge \cite{lin_2014_coco}.
We compute the mAP exactly as in the official COCO evaluation
tool\footnote{\url{https://github.com/cocodataset/cocoapi}} and give its value in
percentage points.
The basic average precision (AP) is the area under the precision-recall curve,
computed for a specific intersection over union (IoU) threshold. In order to
reward algorithms with better localization, the AP is usually averaged over
multiple IoU thresholds, typically the interval $[0.5,0.95]$ in steps of
$0.05$.
The mAP is the mean over APs of all classes in the dataset.  In the following,
we just use the abbreviation AP for the value averaged over IoUs and classes.
When referring to class-averaged AP for a specific IoU threshold, e.g. 0.5, we
write AP$_{50}$.

\subsection{Data Augmentation}
\begin{figure}
	\centering
	\includegraphics[width=0.19\textwidth]{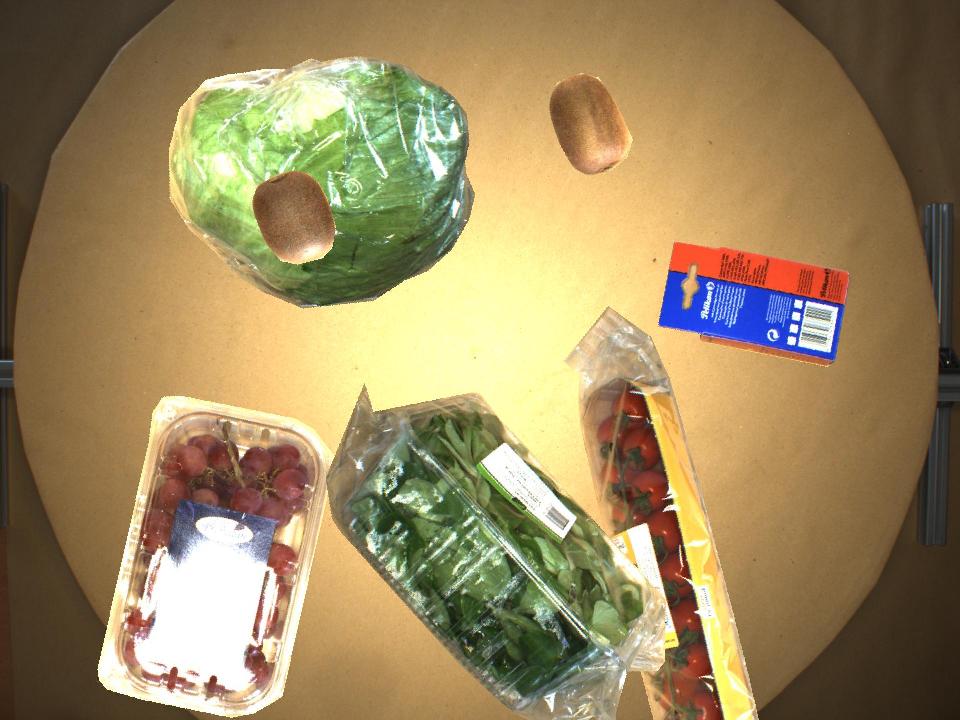}
	\includegraphics[width=0.19\textwidth]{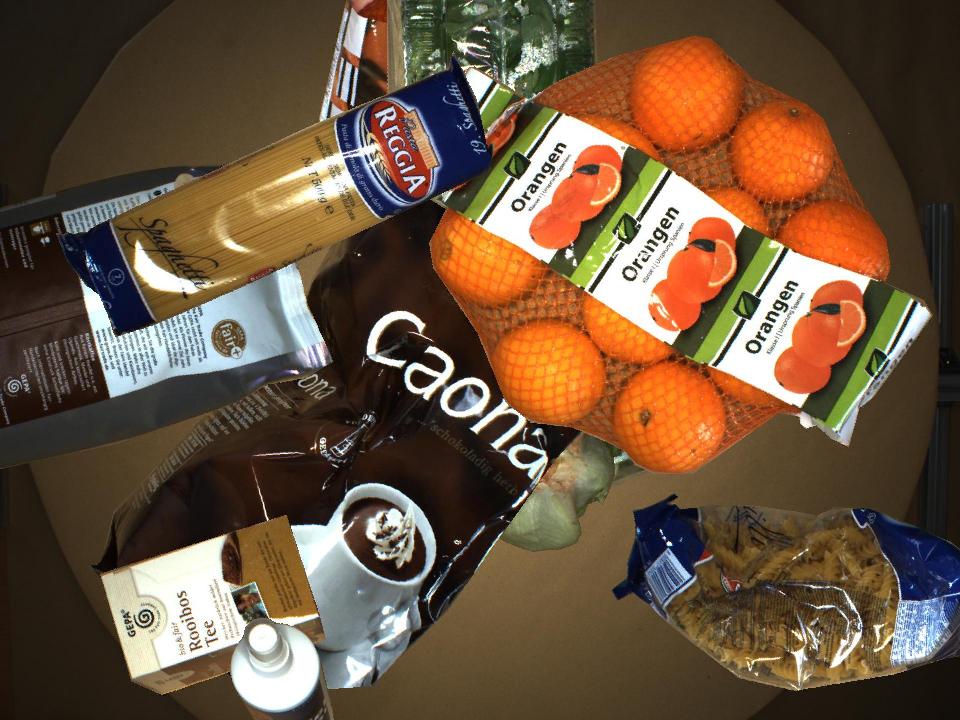}
	\includegraphics[width=0.19\textwidth]{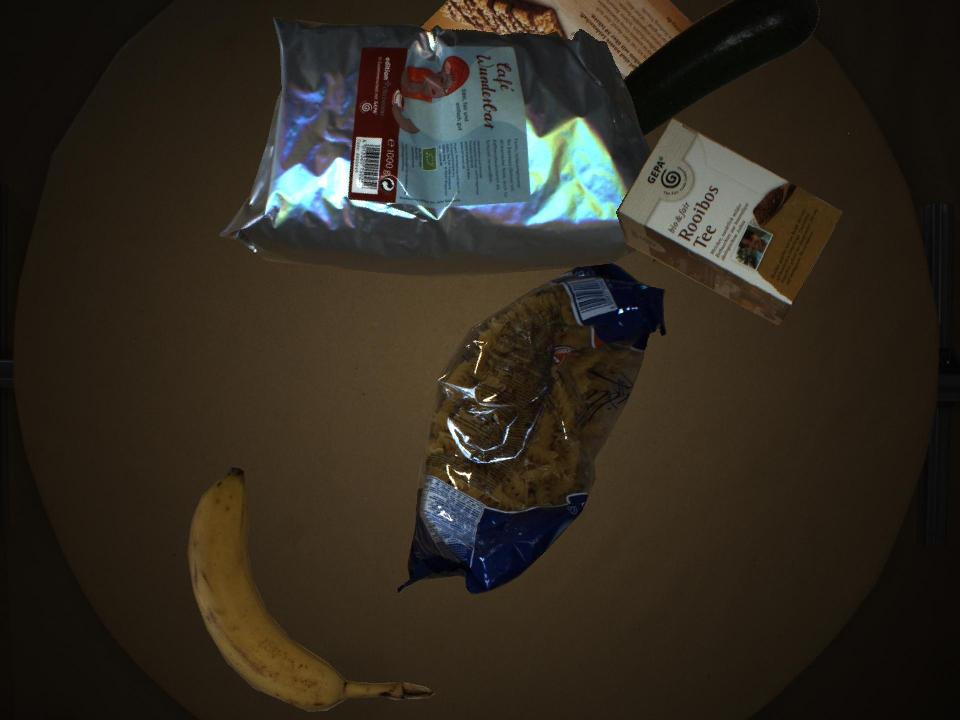}
	\includegraphics[width=0.19\textwidth]{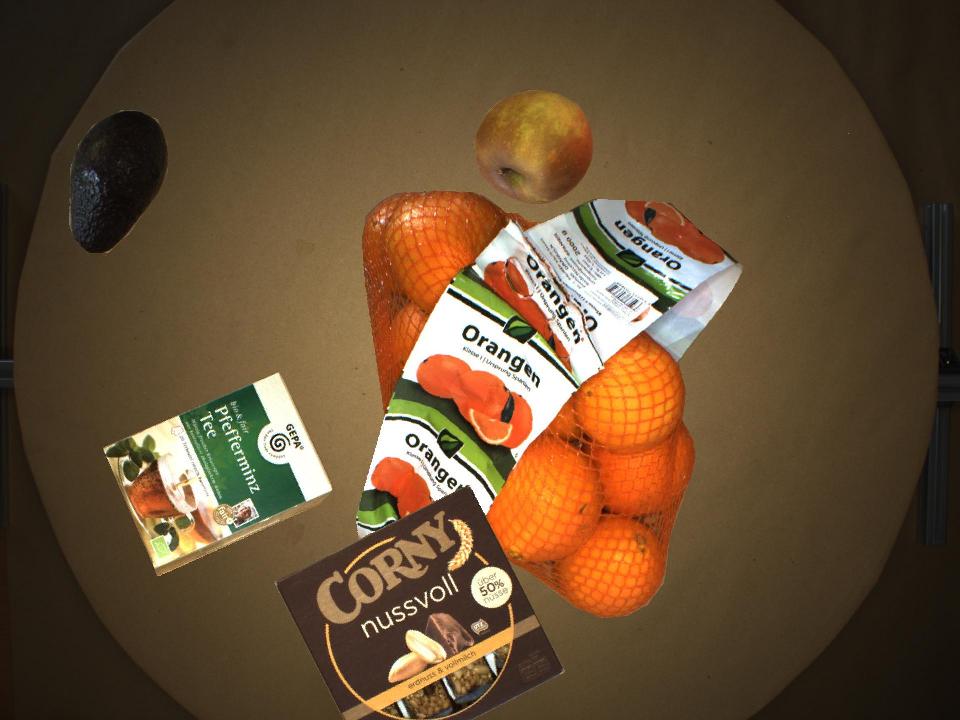}
	\includegraphics[width=0.19\textwidth]{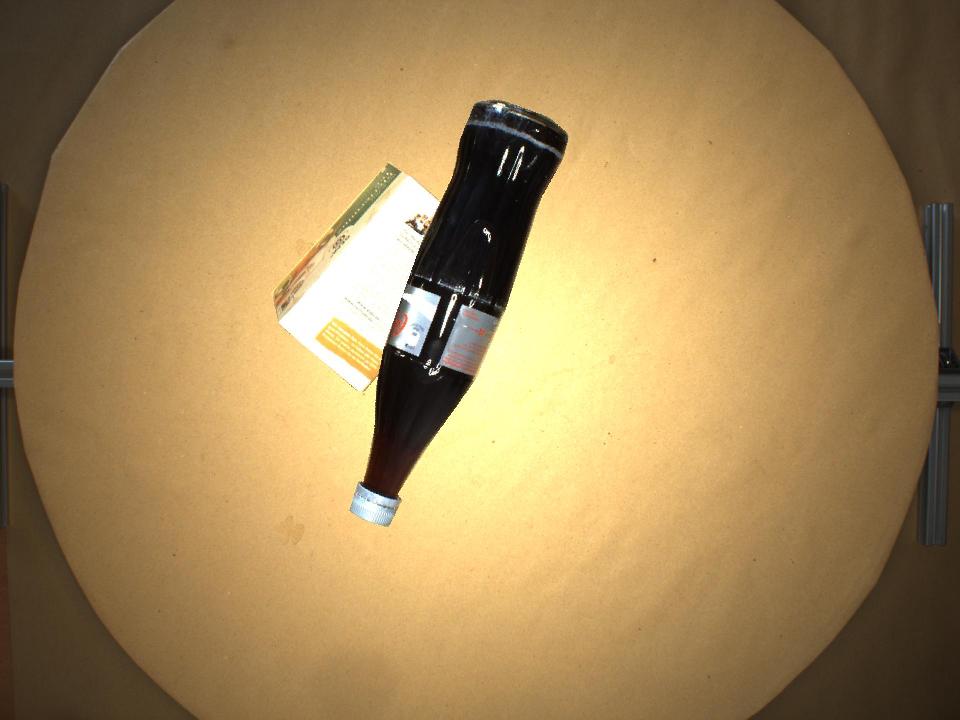}
	\caption{
    The artificial augmented training set is generated by randomly assembling
    objects from the basic training set}
	\label{fig:augmentation}
\end{figure}

In order to keep the labeling effort low and still achieve good results, it is
crucial to artificially augment the existing training set such that it can be
used to train deep neural networks.
Hence, we experiment with a simple data augmentation technique, which serves as
baseline for more sophisticated approaches.  In particular, we simulate the
distribution of validation and test set using only the annotations of the
training set.
For this purpose, we assemble 10\,000 new artificial images that contain one
to fifteen objects randomly picked from the training split.
We denote the augmented data as \texttt{aug} in \tabref{tbl:benchmark_results}.
For each generated image, we randomly sample the lighting and number of object
instances.
For each instance, we randomly sample its class, the orientation, and the
location in the image.
The background of these images is the plain turntable.
We make sure that the instances' region centers lie on the turntable and that
occluded objects have a visible area larger than 5000 pixels.
\figref{fig:augmentation} shows example images of the artificially augmented
dataset for all three different lightings.
Due to the high-quality annotations without margins around the object border,
the artificially assembled images have an appearance that is very similar to
the original test and validation images.

\subsection{Results}
When trained on the full training set \texttt{train} and evaluated on the
\texttt{test} set, the instance segmentation methods provide solid baseline APs
of 49.5\% (Mask R-CNN) and 45.6\% (FCIS).  The object detection results are on
a similar level, with APs of 46.5\% (Mask R-CNN), 44.0\% (FCIS), 46.1\% (Faster
R-CNN), and 51.0\% (RetinaNet).
Tables \ref{tbl:benchmark_results} and \ref{tbl:benchmark_results_bbox} show
the results in full detail.

\paragraph{Ablation Study.}
As aforementioned, the \emph{D2S} splits are based on scenes, i.e. all
rotations and lightings for one placement of objects are included in the same
split. To evaluate the importance of these variations and the ability
of methods to learn invariance with respect to
rotations and illumination, we perform an ablation study.
For this purpose, we create three subsets of the full training set \texttt{train}.
The \texttt{train\_rot0} set contains all three lightings, but only the first
rotation of each scene.
The \texttt{train\_light0} set contains only the default lighting, but all ten
rotations of each scene.
The \texttt{train\_rot0\_light0} set contains only the default lighting and the
first rotation for each scene.

The resulting AP values of the instance segmentation methods Mask R-CNN and
FCIS are summarized in \tabref{tbl:benchmark_results} (top).
As expected, we obtain the best results when training on the full \texttt{train}
set.
Training only on the first rotation reduced the AP on the test set by 15.7\%
and 9.1\% for Mask R-CNN and FCIS, respectively.
Training only with default lighting reduced the AP slightly by
3.4\% for Mask R-CNN and increased the AP by a neglible 0.4\% for FCIS.
Training on \texttt{train\_rot0\_light0} reduced the AP by 13.2\% and 12.9\%,
respectively.
Overall, the results indicate that the models are more invariant to changes in
lighting than to rotations of the objects.

\paragraph{Data Augmentation.}
As shown in \tabref{tbl:benchmark_results}, training on the augmented dataset
\texttt{aug} boosts the AP on the test set to 76.1\% and 69.8\% for Mask R-CNN
and FCIS, respectively.
This is significantly higher than the 49.5\% and 45.6\% achieved by training on
the original \texttt{train} set.
Combining the sets \texttt{train} and \texttt{aug} to \texttt{train+aug}
further improves the AP by 8.3\% and 2.7\%, respectively.


\paragraph{Object Detection.}
We conduct the same ablation study for the task of object detection.
The resulting AP values for all training splits of the methods Faster R-CNN
and RetinaNet, as well as the results of instance segmentation methods Mask
R-CNN and FCIS evaluated on bounding box level, are summarized in
\tabref{tbl:benchmark_results_bbox}. It is interesting to note, that these AP
values are not always better than the AP values obtained for the more difficult
task of instance segmentation.  On the one hand, we believe that the AP values for object detection and instance segmentation are generally very similar because current instance segmentation methods are based on object detection methods like Faster R-CNN. On the other hand, instance segmentation methods can even outperform object detection methods since a nearly perfect mask can still be generated from a too large underlying box proposal. It is also the case that the box-IoU is a lot more sensitive to the four box coordinates than the final segmentation. A third possible explanation is that the gradients of the mask branch help to learn even more descriptive features.
For all methods the overall performance is very similar. Reducing the training
set to only one rotation or only one lighting per scene results in worse
performance. Analogously, augmenting the dataset by generating artificial
training images results in a strong improvement.

\begin{figure}[ht!]
	\centering
	\includegraphics[width=0.19\textwidth]{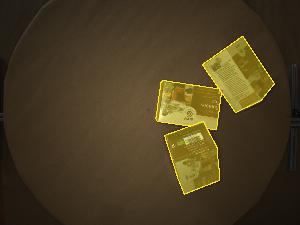}
	\includegraphics[width=0.19\textwidth]{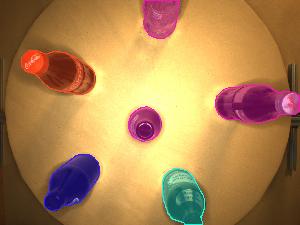}
	\includegraphics[width=0.19\textwidth]{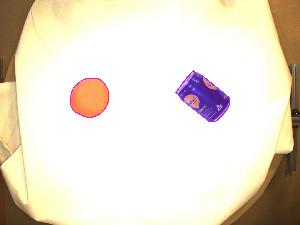}
	\includegraphics[width=0.19\textwidth]{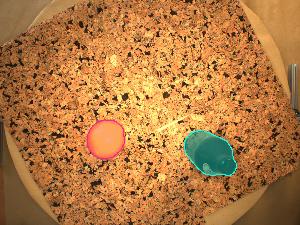}
	\includegraphics[width=0.19\textwidth]{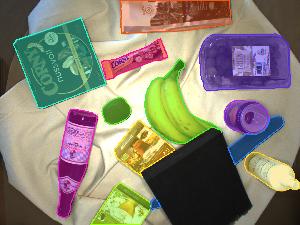}\\
	\includegraphics[width=0.19\textwidth]{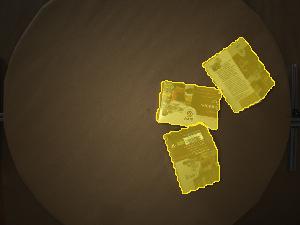}
	\includegraphics[width=0.19\textwidth]{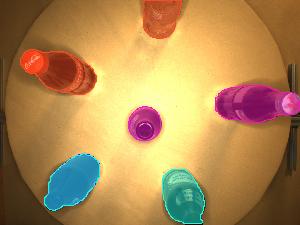}
	\includegraphics[width=0.19\textwidth]{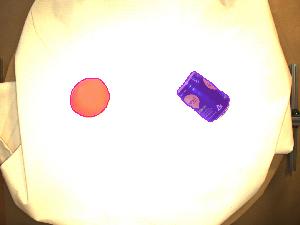}
	\includegraphics[width=0.19\textwidth]{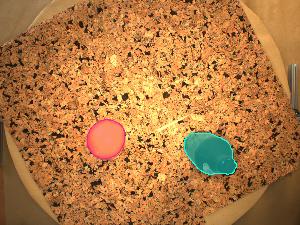}
	\includegraphics[width=0.19\textwidth]{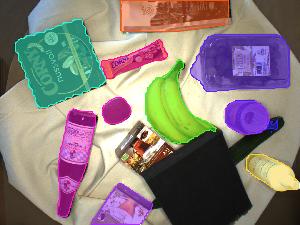}
	\caption{
    (\emph{Top}) Ground truth annotations from the \emph{D2S} \texttt{val} and
    \texttt{test} sets.
    (\emph{Bottom}) Results of Mask R-CNN trained on the \texttt{train} set.
    The classes are indicated by different colors}
	\label{fig:qualitative_results}
\end{figure}

\begin{figure}[t]
	\centering
	\includegraphics[width=0.24\textwidth]{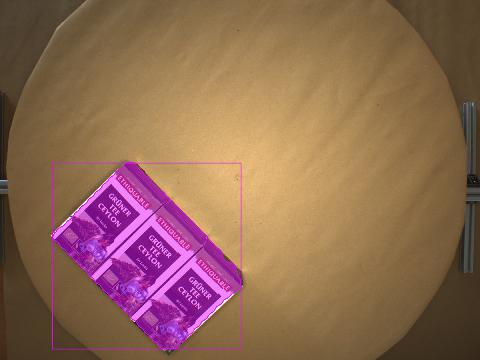}
	\includegraphics[width=0.24\textwidth]{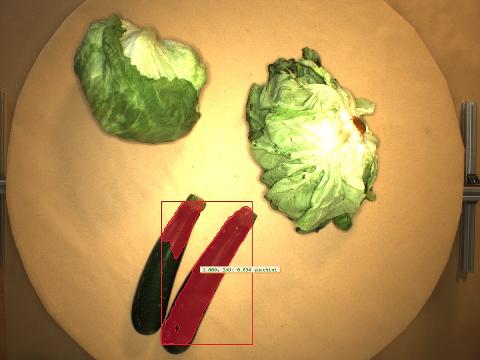}
	\includegraphics[width=0.24\textwidth]{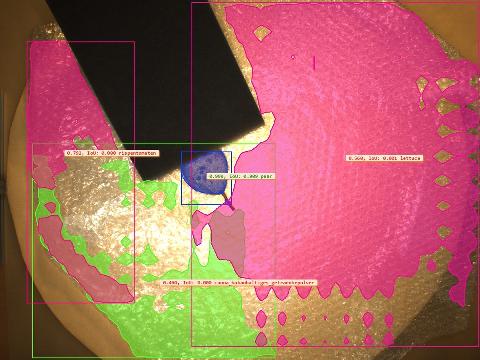}
	\includegraphics[width=0.24\textwidth]{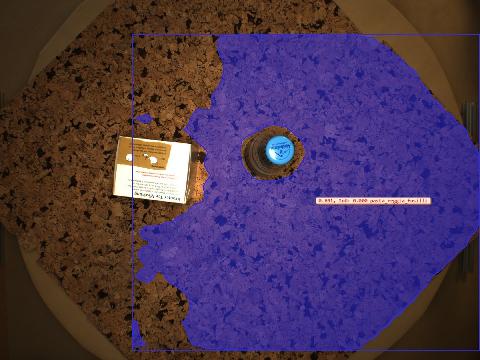}
	\caption{
    Typical failure cases of of Mask R-CNN and FCIS on \emph{D2S}.
    (\emph{From left to right})
    (\emph{1}) Nearby objects are detected as a single instance.
    (\emph{2}) Segmentation mask spans to neighboring objects.
    (\emph{3 and 4}) Background is falsely detected as object
    }
	\label{fig:failure_cases}
\end{figure}

\paragraph{Qualitative results.}
We show qualitative results of the best-performing method Mask R-CNN in
\figref{fig:qualitative_results}.
Furthermore, \figref{fig:failure_cases} shows typical failure cases we
observed for Mask R-CNN and FCIS on the \emph{D2S} dataset.
More qualitative results are provided in the supplementary material.

\begin{table}[t]
  \centering
  \caption{{\bf Instance segmentation benchmark results on the test set.}
    Mean average precision values for models trained on different training
    sets.
    (\emph{Top}) Training on different subsets of the \texttt{train} set.
    (\emph{Bottom}) Training on augmented data yields the highest AP values}
  \label{tbl:benchmark_results}
  \def\arraystretch{1.2}
  \setlength{\tabcolsep}{8pt}
  \begin{tabular}{l|ccc|ccc}
    \hline
        & \multicolumn{3}{c|}{\textbf{Mask R-CNN}} & \multicolumn{3}{c}{\textbf{FCIS}} \\
                              & AP & AP$_{50}$ & AP$_{75}$ & AP & AP$_{50}$ & AP$_{75}$ \\
    \hline\hline
    \texttt{train}               & 49.5  & 57.6  & 51.3  & 45.6  & 58.3  & 51.3 \\
    \texttt{train\_rot0}         & 33.8  & 41.6  & 35.6  & 36.5  & 47.5  & 41.8 \\
    \texttt{train\_light0}       & 46.1  & 54.8  & 48.0  & 46.0  & 59.3  & 52.0 \\
    \texttt{train\_rot0\_light0} & 36.3  & 45.1  & 38.6  & 32.7  & 43.4  & 38.1 \\
    \hline
    \texttt{aug}                 & 71.6  & 86.9  & 81.7  & 69.8  & 87.6  & 82.4 \\
    \texttt{train+aug}           & \textbf{79.9} & \textbf{89.1} &
                                   \textbf{85.3} & \textbf{72.5} &
                                   \textbf{88.1} & \textbf{83.5} \\
    \hline
  \end{tabular}
\end{table}
\begin{table}[t]
  \centering
  \caption{{\bf Object detection benchmark results on the test set.}
    Mean average precision values for models trained on different training
    sets}
  \label{tbl:benchmark_results_bbox}
  \scriptsize
  \def\arraystretch{1.2}
  \setlength{\tabcolsep}{3pt}
  \begin{tabular}{l|ccc|ccc|ccc|ccc}
    \hline
        & \multicolumn{3}{c|}{\textbf{Mask R-CNN}} & \multicolumn{3}{c|}{\textbf{FCIS}} & \multicolumn{3}{c|}{\textbf{Faster R-CNN}} & \multicolumn{3}{c}{\textbf{RetinaNet}} \\
        & AP & AP$_{50}$ & AP$_{75}$ & AP & AP$_{50}$ & AP$_{75}$ & AP & AP$_{50}$ & AP$_{75}$ & AP & AP$_{50}$ & AP$_{75}$ \\
    \hline\hline
    \texttt{train}               & 46.5 & 58.3 & 53.5 & 44.0 & 59.4 & 51.7 & 46.1 & 55.2 & 49.7 & 51.0 & 61.0 & 52.8 \\
    \texttt{train\_rot0}         & 34.1 & 42.5 & 38.3 & 34.6 & 48.2 & 41.3 & 36.7 & 46.9 & 41.5 & 32.9 & 39.8 & 34.5 \\
    \texttt{train\_light0}       & 45.5 & 55.7 & 49.5 & 44.0 & 60.3 & 51.9 & 43.7 & 53.9 & 47.8 & 51.7 & 62.0 & 53.6 \\
    \texttt{train\_rot0\_light0} & 35.7 & 46.0 & 40.5 & 29.9 & 43.9 & 35.4 & 34.3 & 44.3 & 39.0 & 31.6 & 38.9 & 33.2 \\
    \hline
    \texttt{aug}                 & 72.9 & 87.9 & 82.0 & \textbf{69.9} & 88.1 & 80.7 & 73.5 & 88.4 & 82.2 & 74.2 & 86.9 & 81.4 \\
    \texttt{train+aug}     & \textbf{78.3} & \textbf{89.8} & \textbf{84.9}
                           & 68.3 & \textbf{88.5} & \textbf{80.9}
                           & \textbf{78.0} & \textbf{90.3} & \textbf{84.8}
                           & \textbf{80.1} & \textbf{89.6} & \textbf{84.5} \\
    \hline
  \end{tabular}
\end{table}

%% file: section_conclusion.tex
\section{Conclusion}
We have introduced \emph{D2S}, a novel dataset for instance-aware semantic
segmentation that focuses on real-world industrial applications.
The dataset addresses several highly relevant challenges, such as
dealing with very limited training data.  The training set is intentionally
small and simple, while the validation and test sets are much more complex and
diverse.
As opposed to existing datasets, \emph{D2S} has a very uniform distribution of
the samples per class.  Furthermore, the fixed acquisition setup prevents
ambiguities in the labels, which in turn allows flawless algorithms to achieve
an AP of 100\%.
We showed how the high-quality annotations can easily be utilized for
artificial data augmentation to significantly boost the performance of the
evaluated methods from an AP of 49.5\% and 45.6\% to
79.9\% and 72.5\%, respectively.
Overall, the benchmark results indicate a significant room for improvement
of the current state-of-the-art.
We believe the dataset will help to boost research on instance-aware
segmentation and leverage new approaches for artificial data augmentation.